\documentclass{article}

\PassOptionsToPackage{numbers}{natbib}

\usepackage[final]{neurips_data_2023}

\usepackage[utf8]{inputenc} %
\usepackage[T1]{fontenc}    %
\usepackage{hyperref}       %
\usepackage{url}            %
\usepackage{breakurl} 
\usepackage{booktabs}       %
\usepackage{amsfonts}       %
\usepackage{nicefrac}       %
\usepackage{microtype}      %
\usepackage{xcolor}     %
\usepackage{graphicx}
\usepackage{wrapfig}
\usepackage{caption}
\usepackage{float}
\usepackage{fancybox}
\usepackage{listings}

\usepackage{float}
\usepackage{listings}
\usepackage{fancyvrb}
\usepackage{tcolorbox}
\tcbuselibrary{most}
\usepackage{enumitem}
\usepackage{cleveref}
\usepackage{titletoc}
\usepackage{multirow}

\newcommand{\hea}[1]{
\begin{tcolorbox}[enhanced,attach boxed title to top center={yshift=-3mm,yshifttext=-1mm},
colback=blue!5!white,colframe=blue!75!black,colbacktitle=red!80!black,
title=Human Error Analysis, fonttitle = \bfseries, fontupper = \sffamily, fontlower = \sffamily,
boxed title style={size=small,colframe=red!50!black}]
{#1}
\end{tcolorbox}}

\usepackage{booktabs}

\newfloat{lstfloat}{htbp}{lop}
\floatname{lstfloat}{Listing}

\title{ChessGPT: Bridging Policy Learning and \\Language Modeling}

\newtcolorbox{promptbox}{
    fonttitle = \bfseries, fontupper = \sffamily, fontlower = \sffamily,
    title={Prompt}
}
\newtcolorbox{rbox}[1]{
    fonttitle = \bfseries, fontupper = \sffamily, fontlower = \sffamily,
    colbacktitle = red!50!white, title={#1}
}
\newtcolorbox{bbox}[1]{
    fonttitle = \bfseries, fontupper = \sffamily, fontlower = \sffamily,
    colbacktitle = blue!50!white, title={#1}
}
\newtcolorbox{gbox}[1]{
    fonttitle = \bfseries, fontupper = \sffamily, fontlower = \sffamily,
    colbacktitle = teal!50!white, title={#1}
}

\author{%
  Xidong Feng \thanks{Work done during internship at Huawei} \\
  University College London\\
  \And
  Yicheng Luo \\
  University College London \\
  \And
  Ziyan Wang \\
  King's College London \\
  \And
  Hongrui Tang \\
  University College London \\
  \And
  Mengyue Yang \\
  University College London \\
  \And
  Kun Shao \\
  Huawei Noah's Ark Lab\\
  \And
  David Mguni \\
  Huawei Noah's Ark Lab\\
  \And
  Yali Du\\
  King's College London
  \And
  Jun Wang\\
  University College London
}

\begin{document}
\maketitle
\begin{abstract}
When solving decision-making tasks, humans typically depend on information from two key sources: (1) Historical policy data, which provides interaction replay from the environment, and (2) Analytical insights in natural language form, exposing the invaluable thought process or strategic considerations. Despite this, the majority of preceding research focuses on only one source: they either use historical replay exclusively to directly learn policy or value functions, or engaged in language model training utilizing mere language corpus. In this paper, we argue that a powerful autonomous agent should cover both sources. Thus, we propose \textbf{ChessGPT}, a GPT model bridging policy learning and language modeling by integrating data from these two sources in Chess games. Specifically, we build a large-scale game and language dataset related to chess. Leveraging the dataset, we showcase two model examples \textbf{ChessCLIP} and \textbf{ChessGPT}, integrating policy learning and language modeling. Finally, we propose a full evaluation framework for evaluating language model's chess ability. Experimental results validate our model and dataset's effectiveness. We open source our code, model, and dataset at \url{https://github.com/waterhorse1/ChessGPT}.

\end{abstract}
\vspace{-10pt}
\section{Introduction}

In recent years, large language models (LLMs) based on transformer architectures \cite{vaswani2017attention} have showcased remarkable capabilities far exceeding their original design as simple language modeling tools. This was especially notable following the advent of ChatGPT \cite{ouyang2022training}. Stemming from causal language modeling, a plethora of recent studies have concentrated on developing efficient and powerful LLM base models \cite{du2022glm, black2022gpt, touvron2023llama, biderman2023pythia, together2023redpajama}, supervised fine-tuned models \cite{alpaca, vicuna2023, gpt4all, kopf2023openassistant} and models  \cite{ouyang2022training, safe_rlhf, chatglm, moss} leveraging Reinforcement Learning from Human Feedback (RLHF) \citep{christiano2017deep}.
\begin{figure}[t]
\centering
\includegraphics[width=\linewidth]{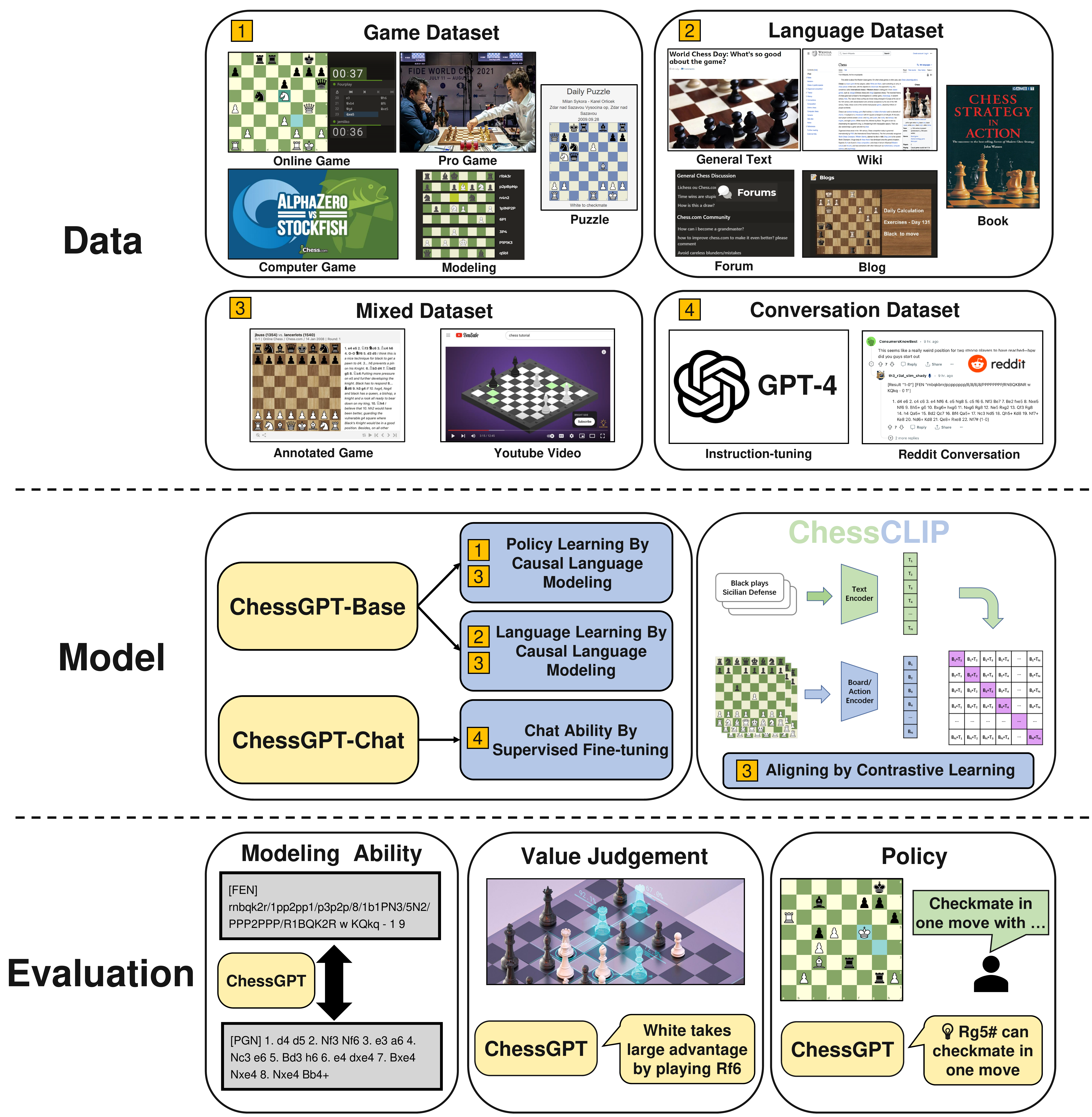}
\caption{Our work provides a comprehensive pipeline that encompasses data, model, and evaluation framework, aiming to foster research on the interaction between policy learning and language learning \textbf{Data}: Our dataset is organized into four categories: game, language, mixed, and conversation datasets. \textbf{Model}: Leveraging this rich dataset, we present two models: ChessGPT and ChessCLIP. \textbf{Evaluation}: Our evaluation framework is structured around three key dimensions: modeling capability, value judgement, and policy proficiency. Refer to~\Cref{apx:image_src} for image sources.}
    \label{fig:main}
    \vspace{-15pt}
\end{figure}

Concurrently, there has been a growing trend in employing Large Language Models (LLMs) as foundational elements for decision-making systems. These systems either depend on the expressive capacity of transformer architectures to execute imitation learning, thereby modeling complex behaviors \cite{chen2021decision, janner2021offline, baker2022video}, or they harness the common knowledge embedded within LLMs to facilitate the policy learning process \cite{wang2023describe, fan2022minedojo, du2023guiding, ahn2022can}. However, the dynamic interplay between policy learning and language modeling has been scarcely addressed. Human decision-making typically involves both: we draw upon historical policy interaction to refine our policy and also employ our thoughts for strategic consideration, mostly in natural language form. Based on this logic, we argue that the study of natural language understanding and policy learning should not be isolated. To advance exploration in this realm, we choose one classic game: \textbf{Chess}, as a practical testbed for initial steps in this direction.

Chess, one of the oldest and most universally played board games, presents an ideal testbed due to the wealth of both policy data and language data. In terms of policy data, it is reported that over ten million games are played daily on Chess.com, the most frequented online chess platform. Regarding language data, a myriad of chess-related knowledge is readily accessible online in various forms and mediums, ranging from game analysis, books, puzzles, and news, to online tutorials, wikis, and even YouTube videos. Building on these resources, we have constructed a comprehensive pipeline dedicated to research on chess policy learning and language modeling, illustrated in fig. \ref{fig:main}. Specifically, we provide the following three components:

\textbf{Datasets.} We curated a large-scale game and language dataset for chess. Our dataset comprises  numerous game data from online chess databases recording how humans and chess engines game replay. It also includes a language dataset that encapsulates chess knowledge in a natural language format, as well as a mixed game-language dataset, which offers the most straightforward interrelated data including articles, discussion, or commentary (language) on specific chess game replay (game).
\textbf{Models.}
We introduce two models, ChessCLIP and ChessGPT, leveraging our datasets. These models showcase the potential for AI to learn from a mixture of replay data and language knowledge.
\textbf{Evaluations.}
We design an extensive set of tasks to evaluate our models' abilities from three distinct perspectives: modeling ability, to gauge the model's proficiency in tracking game state; value judgement ability, measuring  the model's capacity for value assessment and chess knowledge; and policy ability, to test the model's capability in decision-making. Our experimental results confirm that our models consistently outperform other LLM baselines in all evaluation tasks.

We illustrate our full pipeline in fig. \ref{fig:main}. Our work primarily pursues two objectives. Firstly, we construct the whole pipeline on chess as an initial step in promoting research on the interaction/interplay between policy learning and language learning, as well as on the potential of language as a tool for action and understanding. Secondly, our efforts have yielded valuable by-products: the ChessGPT/CLIP models. These models possess practical applicability - they could potentially serve as effective Chess AI assistants for humans.
\section{Related work} 
The pursuit of creating artificial intelligence capable of playing chess can be traced back to the very beginning of the history of computer science~\citep{turing1953digital}. 
Chess engines today achieve superhuman-level performance by utilizing
human knowledge~\citep{campbell2002DeepBlue} or self-play~\citep{silver2018az}.
Recently, there has been increasing interest in improving the interpretability~\citep{mcgrathAcquisitionChessKnowledge2022} of these systems and their alignment with human behavior~\citep{young2020maia} besides strong performance. A chess engine that aligns with human behavior may unlock many exciting opportunities, for example, they can be used as a personalized tutor for chess beginners~\citep{young2020maia}. Some research efforts also concentrated on employing LLMs to learn policies in Chess \citep{noever2020chess, stockl2021watching}. However, these studies mainly center on small-scale datasets or limited training.

There has been increasing interest in leveraging Internet-scale knowledge for creating agents capable of generalizing across many tasks and capabilities \citep{wang2023describe, fan2022minedojo, reed2022generalist}. For example, MineDojo~\citep{fan2022minedojo} introduced a framework on Minecraft for understanding how to enable artificial agents to learn in an open-ended environment. More recently,  there has been a surge in research that treats LLMs as agents, aiming to harness their internet-scale knowledge for decision-making tasks \citep{liu2023agentbench, wang2023voyager, yang2023auto, hong2023metagpt}. In contrast to these studies which typically rely on powerful LLMs like GPT-4 \citep{bubeck2023sparks}, our paper concentrates more on training, especially the interplay between language modeling and policy learning.

\label{gen_inst}
\section{A large-scale game and language dataset for chess}
\label{dataset}
We introduce a large-scale game \& language dataset by collecting chess-related materials from the Internet. Our dataset can be mainly divided into four categories: (1) The Game dataset, encompassing online chess match replay data involving worldwide human players and chess engines of varying skill levels. (2) The Language dataset, principally recording chess-associated knowledge, analyses, discussions, and news in the form of natural language (3) The Mixed Game-Language dataset, incorporating both game data and human natural language elements (such as game analysis or comments) in alignment. (4) The instruction-tuning and conversation dataset, consisting of instruction data and conversation data related to chess. We include comprehensive dataset descriptions, statistics, and text examples in~\Cref{apx:data_stats}, and the procedure of data collection and pre-processing in~\Cref{apx:data}.
\label{headings}
\subsection{Game dataset}
\label{game_dataset}
Game replay data provide the most direct method for both humans and machines to grasp the play mechanics of Chess. In chess, these data are commonly stored in the Portable Game Notation (PGN\footnote{We refer to~\Cref{apx:chess_format} for readers who are not familiar with chess notation format.}) format which is a standard plain text format as illustrated in~\Cref{fig:pgn}.
A PGN starts with some headers that include metadata about the game. These headers include information such as the name of players, the Elo ratings, the opening play, and the game outcome. The headers are followed by a move text section that records the moves played by the two players in turn. The moves may be further annotated with comments enclosed in braces. 

Previous work~\citep{young2020maia} uses the moves recorded in PGNs for policy learning. The moves are interpreted as actions in a Markov Decision Process and the state position can be reconstructed by loading the PGN into a chess engine.
However, PGNs may contain additional useful information beyond the individual moves made. For example, the Elo ratings in the headers may inform us about the relative strength of the players. Additional information included in the comments of the move text section can also be useful - some of the moves are annotated with evaluations generated by computer chess programs that predict the current advantage of the players. These additional annotations may be useful from a reinforcement learning perspective, e.g., for value function learning. For this reason, we curated the game dataset with all of this information intact to facilitate policy learning. 

\begin{wrapfigure}{r}{0.37\linewidth}
\vspace{-0.2in}
\centering
\small
\begin{Verbatim}[frame=single]
[White "Alice"]
[Black "Bob"]
[Result "0-1"]
[WhiteElo "2100"]
[BlackElo "2000"]
[Opening "Sicilian Defense"]

1. e4 { [%
2. Nf3 2... Nc6
...
13. b3?? 13... Nf4? 0-1
\end{Verbatim}
\vspace{-0.05in}
\caption{Replay example in Portable Game Notation (PGN) format.}
\label{fig:pgn}
\vspace{-0.3in}
\end{wrapfigure}
\textbf{Lichess dataset}
We collect five months of online game data from the Lichess database \cite{lichess}, culminating in a total of 17.5 million game replay records for online game players. 

\textbf{Pro-player dataset}
In the Lichess dataset, the majority of player Elo-ratings range between 1000 and 2000. To diversify our game dataset with more skilled matches, we also incorporated an additional 440,000 game records from 245 professional chess players. These professionals typically hold notably higher Elo ratings within the range of 2000 to 2800.

\textbf{CCRL}
Chess engines like StockFish and LeelaZero have attained a proficiency level far beyond what any human player can currently reach. Considering this, we additionally incorporate the \emph{Computer Chess Rating Lists} (CCRL)~\citep{ccrl}, which is a dataset of chess games played by computer chess engines. The CCRL dataset comprises a considerable collection of chess games, specifically 3 million, all of which are played by computer chess engines and stored in PGN format. The Elo-ratings of chess engines fall in the range of 2800-3700.

\textbf{Chess puzzles}
A chess puzzle represents a particular chessboard configuration, designed to present a distinct challenge or objective for the solver. Chess puzzles often require players to find the best move or sequence of moves to achieve a specific goal, such as checkmating the opponent's king, or finding a tactical combination. In our game dataset, we integrate 3.2M puzzles sourced from the Lichess puzzle dataset. Each puzzle within this collection is annotated with its rating, theme and solution.

\textbf{Chess modeling dataset}
We observe that most chess rule descriptions are conveyed in natural language, posing a challenge for machine learning models since they statistically require a large volume of model data to accurately comprehend the chess rules~\citep{schrittwieser2020mastering}. To address this issue, we build a synthetic chess modeling dataset leveraging the python-chess library~\citep{python_chess}. We collect chess game data from a one-month dump of the Lichess dataset, deliberately distinct from the month used in our own Lichess dataset. we design several model-based tasks including converting PGN to FEN, transferring UCI to FEN, and predicting legal moves, etc, resulting in 1.9M data samples.

\subsection{Language dataset}
\label{language_dataset}
\textbf{Existing dataset}
Numerous existing datasets comprise general internet crawl data from platforms like CommonCrawl or Wikipedia. We establish a filtering pipeline to extract only chess-related language corpus from pre-existing language corpus, including C4~\citep{2019t5}, Pile ~\citep{gao2020pile}, Oscar~\citep{OrtizSuarezSagotRomary2019}, Wikipedia~\cite{wikidump} and RedPajama~\cite{together2023redpajama}. These datasets extend the scope of our language data beyond mere game-play.

\textbf{Chess blogs}
Numerous chess websites often publish insightful blogs, sharing their analyses and perspectives on various aspects of chess gameplay. Such blog data is incredibly valuable, as it encompasses game-specific analysis, forming a vital link between the concrete chess game data and its interpretation in natural language form. We manually select approximately 30 chess-related websites and scrape 73.2k blog articles.

\textbf{Chess books}
Similar to chess blogs, chess books can provide long and detailed analysis of the game. We extract approximately 8k chess-related books from online library to enrich our language dataset.

\textbf{Chess forums}
Chess forum serves as a platform for a large amount of chess-related dialogues and conversations involving a diverse range of users. These platforms encompass high-quality question-and-answer pairs, as seen in platforms like StackExchange, or more generalized discussions on various chess-related topics, commonly found in dedicated chess-specific forums. We mainly scrape chess forum data from 5 chess-specific forum platforms and StackExchange, using requests and playwright. This process results in a collection of 140K posts, representing a wealth of diverse views, queries, and discourses related to the world of chess.

\subsection{Mixed game-language dataset}
\label{mixed_dataset}
\textbf{Annotated chess game}
An annotated chess game is a chess game accompanied by written commentary and analysis. In an annotated game, each move made by the players is explained and evaluated, providing insights into the thought process, strategic considerations, and tactical ideas behind the moves. Here is an example of an annotated PGN with Sicilian Defense opening: 

\textit{1.e4 c5
\{The game starts with the Sicilian Defense, one of the most popular and aggressive responses to 1.e4. Black aims to control the center and create imbalances early on.\}}

These annotated games inherently maintain the correspondence between board state and human language, serving as an exceptionally high-quality data source to align a model with complex human intentions and judgements. We amass annotated games from seven sources, five of which are collected from the internet while the rest two are commercial datasets. In total, we collect 245K annotated games with 1.3M board-language pairs.

\textbf{Youtube transcripts}
Drawing inspiration similarly from MineDoJo~\citep{fan2022minedojo}, a YouTube video can naturally serve as a mixed game-language dataset by aligning video clips with natural language transcripts based on timestamps. Rather than generating image-language pairs directly, we develop a pipeline that accurately applies OCR (Optical Character Recognition) to chessboard screenshots to generate FEN (Forsyth-Edwards Notation), a system that describes the chess state in a language format. We gathered around 83k chess videos, resulting in million-scale English transcripts and board-language pairs, thus establishing a substantial mixed game-language dataset.
\subsection{Instruction-tuning \& conversation dataset}
\label{conversation_data}
Supervised fine-tuning is a crucial component to train large language model (LLM) to follow instructions \citep{ouyang2022training, mishra2021cross, wei2021finetuned}. In addition to the comprehensive chess materials mentioned before, we also collect instruction-tuning and conversation datasets which can be used to finetune the pre-trained LLM base model, thereby enhancing its instruction-following and dialogue capability.

\textbf{Instruction-tuning data from GPT-4}
Inspired by Alpaca~\citep{alpaca}, we use the self-instruct technique \citep{selfinstruct}  to generate high-quality, instruction-following data through GPT-4~\cite{bubeck2023sparks}. Specifically, we manually construct 200 seed prompts for chess-related questions or instructions. These prompts serve as few-shot examples, guiding GPT-4 towards more coherent and relevant generation. Finally, we generate around 4k instruction-response pairs using this pipeline.

\textbf{Conversation data from Reddit} The instruction data collected from GPT-4 are mainly in a single-step form, which means only one round of question-answer pair is included. To mitigate this issue, we collect multi-step conversation data about chess on Reddit. Reddit allows users to interact by commenting on posts and responding to other comments, creating a nested structure of responses. This nested structure can be easily converted to a conversation tree by treating the comment's reply as a child node for that reply. A rich source of conversation data can then be acquired by navigating from the root node to each leaf node via every available path. In all, we choose 6 chess-related sub-reddits and collect 410k human conversations about chess.

\section{Large-scale pretraining}
We will showcase two models - \textbf{ChessCLIP} and \textbf{ChessGPT} trained on the large-scale dataset.
\subsection{ChessCLIP}

CLIP (Contrastive Language-Image Pre-Training)~\citep{radford2021clip} is a neural network trained on a variety of modalities (e.g. image, text). By conducting contrastive learning on a large amount of paired data, CLIP bridges the image and language modality, enabling the model to understand vision by language information and vice versa. Our mixed game-language dataset in~\Cref{mixed_dataset} has a similar paired structure because the annotation is naturally paired with its preceding game trajectories. Based on this subset, we can train a \textbf{ChessCLIP} to bridge the modality of policy and language. Specifically, by denoting the chessboard state $S$ at timestep $t$ as $S_{t}$, and the annotation language as $L_{t}$, the data pair at timestep $T$ can be represented by $\big((\{S_{t}\}_{t=T-k}^{t=T}, a_{T}), L_{T}\big)$ where $\{S_{t}\}_{t=T-k}^{t=T}$ is a stacked $k$ history states and $a_T$ is the last move.

We want to emphasize more on what ChessCLIP can do by aligning the policy modality and the language modality. Firstly, ChessCLIP offers a similarity metric given one PGN and a text description. Just like the application of large-scale image/text retrieval using CLIP, ChessCLIP can help users conduct PGN/text retrieval - search for game based on text or search for comments based on specific game. In addition, because of the low-dimensional feature of action space compared to vision or language space (there only exists a few legal moves for a given chess state ), we can directly conduct search algorithms to maximize the similarity to generate action based on one text description using ChessCLIP. For example, given a chessboard state and a text description, ChessCLIP can generate a move by iterating through all legal moves and finding one move that returns the largest similarity. By the same logic, ChessCLIP can directly generate move sequences (multiple actions) using greedy search or beam search. We refer the reader to~\Cref{apx:chessclip_details} for more discussions.
 
\textbf{Implementation details} 
We preprocess the annotated PGNs to produce board/text pairs which we feed separately to the board and text encoders. In particular, for every move in the PGN, we extract the comments attached to the move as well as the board state. We encode the board positions and moves using the same scheme as those used by Leela Chess Zero (lc0)~\citep{LeelaChessZero}, which is similar to the encoding used by AlphaZero~\citep{silver2018az} for encoding positions and moves in chess. Concretely, the board positions are encoded as a $\mathcal{R}^{8 \times 8 \times 112}$ feature map and the actions are encoded as a $\mathcal{R}^{1858}$ vector. We instantiate a ChessCLIP model with a pair of text encoder and a board/action encoder. For the text encoder, we only fine-tune the last two layers of pretrained text encoder from OpenAI CLIP model. For the board/action encoder, we use a ResNet~\citep{he2016resnet} architecture that conditions on the action encoding via a modified FiLM layer~\citep{perez2018film}.
Please refer to~\Cref{apx:chessclip_details} for implementation details.
\subsection{ChessGPT}
The Generative Pretraining Transformer (GPT-3)~\citep{brown2020language} is an autoregressive language model that uses deep learning techniques to generate human-like text. GPT-3 is trained by casual language modeling, which aims to predict the next word in a sentence given all the previous words. Following the same logic, we train a GPT-like model using all chess materials introduced in~\Cref{dataset}. Unlike other policy behavior data in robots ~\citep{levine2016end} or video games~\citep{mnih2013playing}, the chess state and move data can be represented in merely textual format. Thanks to this feature, we can directly treat chess as a text game and the imitation learning objective for policy learning can be directly covered by casual language modeling over the game dataset provided in~\Cref{game_dataset}.

\textbf{Implementation details}
We follow common implementations of training a domain-specific instruction-following LLM. Firstly we conduct base-model fine-tuning using chess corpus introduced in section~\ref{game_dataset},~\ref{language_dataset} and ~\ref{mixed_dataset}. Due to computational constraints, we choose to finetune the RedPajama-3B-base~\cite{together2023redpajama} model, which is an open-souce replication of LLaMA~\cite{touvron2023llama}. The base model adopts the GPT-NeoX~\cite{black2022gpt} architecture, a GPT-3~\citep{brown2020language} variant with a few modifications such as rotary positional embedding, parallel attention computation, and different initialization. The base-finetuning brings us our base model: \textbf{ChessGPT-Base}. After base-finetuning, we conduct supervised fine-tuning by supervised learning on question/conversation response using data introduced in section~\ref{conversation_data} and general conversation data from OASST1~\citep{kopf2023openassistant}, Dolly2 ~\citep{dolly2}, Alpaca-GPT4~\citep{peng2023instruction}, and Sharegpt ~\citep{sharegpt}, forming our chat model: \textbf{ChessGPT-Chat}.
We leave further RLHF (Reinforcement Learning from Human Feedback) training for future work.
Refer to~\Cref{apx:chessgpt_details} for more details.
\label{others}

\section{Evaluation and benchmark}
In this section, we present a comparative analysis between ChessGPT trained on our database with other baseline LLMs. The purpose of our experiments is to assess the performance of ChessGPT in three primary dimensions: Chess modeling ability, Value judgement ability, and Policy ability. The Chess Modeling capability focuses on the language model's proficiency in accurately tracking the game state and predicting valid moves. Regarding the Value judgement ability, we assess the model's precision in evaluating the worth of a chess game, encompassing the identification of advantageous positions and the calculation of situation scores. Lastly, the Policy capability gauges the model's aptitude for generating optimal moves based on a given position. By thoroughly examining these sub-categories, we can comprehensively evaluate and contrast the efficacy of different models in chess-related tasks. We choose the following models as baselines: LLaMA-7B~\citep{touvron2023llama}, RedPajama-Base-3B~\citep{together2023redpajama}, and compare them with  ChessCLIP, ChessGPT-Base-3B\footnote{The model size is 2.8B and 3B is a rounded representation}, and ChessGPT-Chat-3B. To help readers who are not familiar with chess, we provide task examples and illustrative figures to elucidate these evaluation tasks in~\Cref{apx:eval_figs}.

\subsection{Chess modeling ability}
\label{modeling}
\textbf{Chess state tracking} We utilized Big-bench's State Tracking in Chess task~\citep{srivastava2022beyond, toshniwal2022chess} to evaluate language models' ability to track the state of chess games encoded in UCI notation. The task involves predicting the legal ending square given the game prefix and starting square of the current move. For example, if the input UCI notation is "$f2f4\:\:\:d7d5 \:\:\: g1$", the expected output would be $["h3","f3"]$, as the chess piece on square $g1$ can only move to those two positions. The task dataset includes real and synthetic games, divided into short, medium, and long categories based on move count. The evaluation measures correctness across all games using a specified output regex. Notably, the ChessCLIP is unsuitable for modeling tasks, so we do not include it in the comparison.

\Cref{tab:stc} presents a performance analysis of all models on the task. Our Base and Chat models consistently outperformed baselines in all tasks. This indicates their strong ability to track the state of chess games. However, the ChessGPT-Chat model exhibited slightly lower performance, suggesting a potential trade-off between language capabilities and state tracking. Nevertheless, the results underscore the effectiveness of our dataset-trained LLM models for chess state tracking.

\begin{table}
\caption{Bigbench State Tracking in Chess}
\small
\centering
\begin{tabular}{ccccc}
\toprule
& \multicolumn{4}{c}{LLM Models (\%)} \\
\cmidrule{2-5} 
Tasks      & LLAMA-7B & RedPajama-Base & ChessGPT-Base & ChessGPT-Chat \\
\midrule
Real Short &   29.5 $\pm$ 1.4&      {23.2 $\pm$ 1.3}            &   \textbf{99.5 $\pm$ 0.2}                   &     \textbf{98.5 $\pm$ 0.4}                                          \\
Real Med   &    39.3 $\pm$ 1.5     &      {38.2 $\pm$ 1.5}          &    \textbf{97.7 $\pm$ 0.5}                  &    \textbf{97.8 $\pm$ 0.4}                                           \\
Real Long  &    53.0  $\pm$ 1.6    &    {51.9 $\pm$ 1.6}     & \textbf{98.1 $\pm$ 0.4}  &   \textbf{97.6 $\pm$ 0.4}                                            \\
Syn Short  &   31.3  $\pm$ 1.4     &      24.9  $\pm$ 1.3       &     \textbf{94.2 $\pm$ 0.7}               &         \textbf{92.3 $\pm$ 0.8}                                  \\
Syn Med    &   39.9  $\pm$ 1.6     &    37.7  $\pm$ 1.5           &  \textbf{94.6 $\pm$ 0.7}                    &            \textbf{88.9 $\pm$ 1.0}                               \\
Syn Long   &  45.8 $\pm$ 1.5      &  42.2  $\pm$ 1.5           &     \textbf{92.8 $\pm$ 0.8}             & \textbf{85.1 $\pm$ 1.1}            \\
\bottomrule
\end{tabular}
\label{tab:stc}
 \vspace{-0.2in}

\end{table}
\textbf{Board state tracking} We performed additional evaluations involving UCI to FEN and PGN to FEN conversions. In the UCI to FEN experiment, the target was replaced with FEN format, while in the PGN to FEN experiment, UCI was converted to PGN format as input and the target was replaced with FEN format. The similarity was measured using Levenshtein distance, which was normalized to a range of 0 to 1~\citep{yujian2007normalized}. These evaluations focused on assessing the model's capability to track the overall state of the chessboard by representing the state of each chess piece using FEN notation.
\begin{table}
\caption{UCI to FEN test}
\small
\centering
\begin{tabular}{cccccc}
\toprule
& & \multicolumn{4}{c}{LLM Models (\%)} \\
\cmidrule{3-6} 
& Tasks      & LLAMA & RedPajama-Base & ChessGPT-Base  & ChessGPT-Chat \\
\midrule
& Real Short &  2.2 $\pm$ 0.0 &   5.2 $\pm$ 0.0 &          \textbf{95.1  $\pm$ 0.3}          &  \textbf{95.7 $\pm$ 0.1}  \\

UCI to FEN & Real Med  & 2.3 $\pm$ 0.1 &  4.0 $\pm$ 0.1       &      \textbf{89.9   $\pm$ 0.2}            &    \textbf{88.6 $\pm$ 0.3}  \\
& Real Long  & 1.8 $\pm$ 0.0  &  3.8     $\pm$  0.1    &    \textbf{85.7 $\pm$ 0.2}        &  \textbf{81.4 $\pm$ 0.5}   \\ 
\midrule
& Real Short & 6.0 $\pm$ 0.1 &   2.8 $\pm$  0.1   &  \textbf{96.7  $\pm$ 0.1 }    &  \textbf{95.8 $\pm$ 0.1} \\

PGN to FEN & Real Med  & 5.6 $\pm$ 0.1  & 3.4 $\pm$ 0.1    &     \textbf{94.8 $\pm$ 0.1}        &   \textbf{93.6  $\pm$ 0.1} \\ 
& Real Long  & 5.7 $\pm$ 0.0  &   8.9 $\pm$ 0.1 &     \textbf{95.4  $\pm$ 0.2}   &  \textbf{82.7 $\pm$ 1.0} \\
\bottomrule
\end{tabular}
\label{tab:u2f}
 \vspace{-0.2in}
\end{table}

\Cref{tab:u2f} illustrates the results of these evaluations. It is evident that compared to tracking the state of an individual chess piece, tracking the entire chessboard state becomes more challenging. The similarity scores between the two baselines were consistently below $10\%$, indicating a lack of global chess piece state tracking ability. However, the ChessGPT achieves an average similarity score higher than $90\%$. These results demonstrate that our dataset-trained model excels in capturing and reproducing the global chess piece state in both UCI to FEN and PGN to FEN conversions.

\subsection{Value judgement ability}
In this part, we evaluate the model's ability of value judgement. Specifically, we want to assess the model from two perspectives: (1) its ability to align with the true value function given a chessboard state (the true value are evaluated by chess engines in enough search depths) in the evaluation of \textbf{State value multi-choice}, and (2) its ability to align with human judgement and human knowledge in the evaluation of \textbf{Chess Annotation Multi-choice} and \textbf{Opening multi-choice}.

\textbf{State value multi-choice} 
Here we evaluate the model's ability to see whether it can determine which side holds the advantage for a given PGN. We construct an evaluation dataset consisting of $3000$ game snippets and utilize Stockfish 15 with a depth of 18 to calculate the winning rate for the white pieces. By categorizing the winning rate into three intervals: $0-33\%$ for black advantage, $34-66\%$ representing a balanced state, and $67-100\%$ for white advantage, we construct the state-value multiple-choice task. During experiments, we discovered that an additional '\{' suffix to the prompt can significantly enhance the performance of the base model. This is due to '\{' consistently serving as the initial symbol for annotation in annotated PGNs. Consequently, we carried out our evaluation under two distinct prompt settings and report our results w.r.t multi-choice grade shown in~\Cref{tab_state_value}.
\begin{table}[t]
\caption{State value multi-choice}
\label{tab_state_value}
\small
\centering
\begin{tabular}{cccccc}
\toprule
& \multicolumn{4}{c}{Models (\%)}                                                                 \\ \cmidrule{2-6} 
Prompt Setting & LLAMA & RedPajama & ChessGPT-Base & ChessGPT-Chat  & ChessCLIP \\ \midrule
W/O $\{$ suffix & {33.2 $\pm$ 0.7}  &      {31.1 $\pm$ 0.7}            &   \textbf{43.1 $\pm$ 0.8}         & \textbf{52.8 $\pm$ 0.8}   &  N/A                                          \\
With $\{$ suffix   &    {26.9 $\pm$ 0.7}     &      {29.7 $\pm$ 0.8}          &  \textbf{53.7 $\pm$ 0.8}   & \textbf{53.5 $\pm$ 0.8}  & \textbf{38.1 $\pm$ 0.8}                                      \\ 
\bottomrule
\end{tabular}
\vspace{-0.2in}
\end{table}

\textbf{Chess annotation multi-choice}
The annotations within an annotated PGN can be viewed as a reflection of human evaluative judgement. To examine the degree to which the model's value aligns with human value, we extract 3k game-language pairs from the annotation dataset as the test set. By randomly selecting three annotations from the test set as incorrect options, we construct the chess annotation four-choice task. We report the multi-choice grade results over two prompts in Table \ref{tab_annoation}.
\begin{table}[t]
\caption{Chess Annotation Multi-choice}
\label{tab_annoation}
\small
\centering
\begin{tabular}{cccccc}
\midrule
& \multicolumn{4}{c}{Models (\%)} \\ 
\cmidrule{2-6}
Prompt Setting & LLAMA & RedPajama & ChessGPT-Base & ChessGPT-Chat  & ChessCLIP \\ 
\midrule
W/O $\{$ suffix &   {29.8 $\pm$ 0.8}  &      {27.4 $\pm$ 0.7}            &   \textbf{33.2 $\pm$ 0.9}  & \textbf{35.7 $\pm$ 0.9} & N/A                                                \\
With $\{$ suffix   &    {29.6 $\pm$ 0.8}       &      {28.4 $\pm$ 0.8}          &  38.8 $\pm$ 0.9   & {34.7 $\pm$ 0.9} &   \textbf{63.6 $\pm$ 0.9}                                        \\ 
\bottomrule
\end{tabular}
\vspace{-0.2in}
\end{table}

\textbf{Opening multi-choice}
A chess opening refers to the initial moves made by players at the beginning of a chess game. There are numerous chess openings, each with its own name, characteristics, and strategic goals. For example, the Sicilian defense: \textit{1. e4 c5} is one of the most popular and aggressive chess openings for Black. We use the Lichess opening dataset \citep{opening} including 3.5k opening PGNs and their corresponding names, to formulate two tasks: (1) PGN2Opening five-choice task, which aims at choosing the correct opening name for a given PGN, and reversely, (2) Opening2PGN five-choice task, aiming at choosing the correct PGN for a given opening name. We report the result in~\Cref{tab_opening}.

In general, our trio of models surpasses the performance of two baseline language models across these four tasks in all settings. This result confirms that our models are more effectively aligned with both the true value function and human judgement/knowledge. Both ChessGPT-Base and ChessGPT-chat deliver outstanding performance in the state-value task and the opening task. Notably, ChessCLIP displays a surprisingly high level of proficiency in the annotation task and the opening task. This result reveals the model's capacity to extract human judgement and knowledge solely from annotations, even without training in any actual chess games.

\begin{table}[t]
\caption{Opening2PGN and PGN2Opening}
\label{tab_opening}
\small
\centering
\begin{tabular}{cccccc}
\toprule
& \multicolumn{4}{c}{Models (\%)} \\ 
\cmidrule{2-6} 
Prompt Setting & LLAMA & RedPajama & ChessGPT-Base & ChessGPT-Chat  & ChessCLIP \\ 
\midrule
Opening2PGN & {43.0 $\pm$ 0.9}  &      {26.5 $\pm$ 0.8}            &   \textbf{92.2 $\pm$ 0.5}         & \textbf{94.7 $\pm$ 0.4}    &  {73.0 $\pm$ 0.8}                                              \\
PGN2Opening   &   {20.0 $\pm$ 0.7}      &      {20.7 $\pm$ 0.7}          &  49.3 $\pm$ 0.9   &  {55.8 $\pm$ 0.9}  & \textbf{80.5 $\pm$ 0.7}                                      \\ 
\bottomrule
\end{tabular}
\vspace{-0.2in}
\end{table}

\begin{table}[t]
\caption{Checkmate in One}
\label{tab_checkmate}
\small
\centering
\begin{tabular}{cccccc}
\toprule 
& \multicolumn{4}{c}{Models (\%)} \\
\cmidrule{2-6}
Setting      & LLAMA & RedPajama  & ChessGPT-Base & ChessGPT-Chat & ChessCLIP \\ 
\midrule
With suffix (ESM) &  {1.6 $\pm$ 0.2} &      {0.0 $\pm$ 0.0}  &    \textbf{71.4 $\pm$ 0.7}                &    {56.8 $\pm$ 0.8}    &         N/A  \\
With suffix (MC) & {2.6 $\pm$ 0.3} & {0.0 $\pm$ 0.0} & \textbf{66.1 $\pm$ 0.8} & {11.3 $\pm$ 0.5} &    {2.9 $\pm$ 0.3} \\
W/O suffix (ESM)  &   {1.7 $\pm$ 0.2}      &      {0.0 $\pm$ 0.0}          & {26.5 $\pm$ 0.8} & \textbf{59.4 $\pm$ 0.8}  &    N/A \\
W/O suffix (MC)   & {2.2 $\pm$ 0.3}    & {0.0 $\pm$ 0.0} &  {13.6 $\pm$ 0.6}    &   \textbf{15.4 $\pm$ 0.6}   &   N/A \\
\bottomrule
\end{tabular}

\end{table}
\subsection{Policy evaluation} 
\textbf{Checkmate in one} We incorporate the checkmate-in-one task from Big-Bench \cite{srivastava2022beyond} into our evaluation methods. This task is designed to challenge the model's ability to identify a move in a given PGN that would result in a checkmate. By doing so, it measures the model's capacity to comprehend and apply the rules of chess. The model is essentially required to discern a move that not only places the opponent's king under attack but also ensures that the king cannot evade capture in the next move.

We also find adding an additional instruction suffix like \textit{\{Now white/black can checkmate in one\}} can largely enhance the base model performance. We report the result in two prompts with two metrics (exact-string match as ESM and multi-choice-grade as MC) in Table \ref{tab_checkmate}. our ChessGPT-Base model and ChessGPT-Chat model show a really great checkmate ability by surpassing two LLM baselines by a large margin. ChessCLIP does not perform well in this task, because there does not exist much annotation data regarding checkmate-in-one behavior in the annotation dataset.
\begin{wrapfigure}{r}{0.33\textwidth}
    \vspace{-0.15in}
    \centering
    \begin{minipage}{0.32\textwidth}
    \begin{table}[H]
    \centering
    \caption{Elo Rating 1700-2000}
    \small
    \begin{tabular}{ccc}
      \toprule
      LLM Models    &  Move Score  \\
      \midrule
      LLAMA &  {55.1 $\pm$ 1.1}\\
      RedPajama & {56.4 $\pm$ 0.9} \\
      ChessGPT-Base & {59.6 $\pm$ 1.0} \\
      ChessGPT-Chat & {60.3 $\pm$ 1.0} \\
      \bottomrule
    \end{tabular}
    \label{tab:elorating1700}
    \end{table}
\end{minipage}
\vspace{-0.2in}
\end{wrapfigure}

\textbf{General policy} In order to assess the model's generalization ability, we introduced Elo Rating as a factor in the task, aiming to evaluate its capacity to identify PGN and related keywords and generate the appropriate next move within the specified skill level. Model's selection of the next legal move is assigned a move score, which is normalized based on the win rate observed in the raw data. \Cref{tab:elorating1700} presents the results representing the performance of different models in selecting the most suitable move for white chess. Notably, all models surpassed the performance of the random policy ($\approx 50\%$) as the Elo Ratings correspond to relatively high skill levels among human players.

Further analyzing the performance of different models across varying Elo Ratings is crucial for understanding the observed results. The minor variations in move scores for different Elo Rating scenarios in~\Cref{tab:chassgpt-base-gen-test} indicate that ChessGPT-Base may struggle to effectively incorporate Elo Rating information into its decision-making process. This could be due to the model's limited understanding of the nuanced characteristics associated with distinct Elo Ratings. The complexity of the task and the challenges in accurately accounting for diverse playing styles further contribute to the limited variations in move scores across different Elo Ratings. Consequently, neglecting this information can lead to the model learning an average policy for each Elo Rating, resulting in subpar overall performance. Similar findings were observed in the black chess test, and to further validate this viewpoint, we conducted input attention visualization. Refer to~\Cref{apx:chessgpt_details} for more details.

\begin{wrapfigure}{r}{0.3\textwidth}
        \vspace{-0.4in}
        
    \begin{minipage}{0.29\textwidth}
    \begin{table}[H]
    \centering
    \small
    \caption{ChessGPT-Base in Different Elo Rating Results}
    \begin{tabular}{cc}
      \toprule
      Elo Rating    & Move Score \\
      \midrule
      700-1000 &  {59.4 $\pm$ 1.0} \\
      1200-1500 & {58.9 $\pm$ 0.9} \\
      1700-2000 & {59.6 $\pm$ 1.0} \\
      2700-3000 & {59.8 $\pm$ 1.0} \\
      \bottomrule
    \end{tabular}
    \label{tab:chassgpt-base-gen-test}
     \end{table}
\end{minipage}
    \vspace{-0.15in}
\end{wrapfigure}

To clarify, the dataset we have presented encompasses a wide range of games and varying Elo ratings, as shown in Figure \ref{fig:pgn}, which possesses the potential to effectively capture and generalize intricate patterns and policies associated with different Elo levels. However, the current training method might not sufficiently emphasize these nuanced features. This highlights a potential direction for future research, which involves enhancing the model's ability to better integrate and utilize metadata such as Elo Rating and other auxiliary data. By addressing these aspects, the model's overall generalization can be further improved.

\subsection{Qualitative results}
We also perform qualitative comparison between our models (ChessGPT-Chat and ChessGPT-Base) and the baselines. We ask the language models a series of questions ranging from factual knowledge of chess as well as requesting the models to perform some operational tasks related to chess. We found that ChessGPT-base performed similarly to RedPajama: both models can sometimes produce factual answers for some of the questions but they failed to generate coherent answers when asked to perform tasks such as providing commentary on chess moves or converting the PGN notation to FEN. ChessGPT-Chat gives more factual answers and demonstrates better performance when prompted to generate analysis and perform other chess-related tasks. Refer to  \Cref{apx:qualitative-analysis} for qualitative analysis.

\section{Conclusion}
In this paper, we introduce a new large-scale dataset and benchmark on chess to encourage of study of the interplay between historical policy data and natural language knowledge. We accompany our dataset with an evaluation framework for assessing language models' capability in chess. We showcase two models, \textbf{ChessCLIP} and \textbf{ChessGPT}, that demonstrate promising results for learning the interplay between language and action. Nevertheless, our results indicate that we are only beginning to understand how to bridge the gap between policy learning and language modeling and we discuss more on the future directions of our dataset in~\Cref{apx:potential-directions}.
We hope that our dataset and benchmark can make future research on policy and language alignment more accessible.

\clearpage
\bibliographystyle{plain}
\bibliography{cite}

\newpage
\startcontents[appendix]
\appendix

\section*{Appendix}

\renewcommand*\contentsname{Appendix}

\printcontents[appendix]{ }{1}{\setcounter{tocdepth}{2}}
\clearpage

\section{Acknowledgement}
We thank Jiacheng Liu for his work on collecting chess-related data and chess book list.

\section{Different Chess Formats}
\label{apx:chess_format}

\subsection{Universal Chess Interface (UCI)}
The UCI format is widely used for communication between chess engines and user interfaces. It represents chess moves by combining the starting and ending squares of a piece, such as "e2e4" to indicate moving the pawn from e2 to e4. For example, the UCI notation for a full game would be:

\small
\begin{Verbatim}[frame=single]
e2e4 c7c6 g1f3 d7d5 e4d5 c6d5 d2d4 b8c6 c2c4 g8f6 b1c3 c8e6 c4c5 g7g6 c1f4 f8g7 
f1e2 f6e4 e1g1 e6g4 f3e5 g4e2 d1e2 c6e5 c3e4 e5c6 e4d6 e8f8 f1e1 g7d4 f4h6 d4g7
h6g7 f8g7 d6b5 a7a6 b5c3 d5d4 c3e4 d8d5 a2a3 a8d8 b2b4 h7h6 e2
\end{Verbatim}

\subsection{Standard Algebraic Notation (SAN)}
SAN (Standard Algebraic Notation) is a widely used notation system in the game of chess for recording and describing moves. It provides a standardized and concise representation of moves that is easily understood by chess players and enthusiasts. In SAN, each move is represented by two components: the piece abbreviation and the destination square. The piece abbreviation is a letter that represents the type of piece making the move, such as "K" for king, "Q" for queen, "R" for rook, "B" for bishop, "N" for knight, and no abbreviation for pawns. The destination square is denoted by a combination of a letter (a-h) representing the column and a number (1-8) representing the row on the chessboard. Additional symbols may be used to indicate specific move types. The symbol "+" is used to indicate a check, while "\#" denotes a checkmate. Castling moves are represented by "O-O" for kingside castling and "O-O-O" for queenside castling.

\subsection{Portable Game Notation (PGN)}
PGN is a widely adopted format for recording chess games. It includes not only the SAN moves but also additional information like player names, event details, and game results. PGN files are human-readable and can be easily shared and analyzed. Here is an example of a PGN representation of a full game:

\small\begin{Verbatim}[frame=single]
[Event "World Chess Championship"]
[Site "London, England"]
[Date "2023.05.20"]
[Round "1"]
[White "Carlsen, Magnus"]
[Black "Nepomniachtchi, Ian"]
[Result "1/2-1/2"]

1. e4 e5 2. Nf3 Nc6 3. Bb5 a6 4. Ba4 Nf6 5. O-O Be7 6. Re1 b5 7. Bb3 d6 
8. c3 O-O 9. h3 Nb8 10. d4 Nbd7 11. Nbd2 Bb7 12. Bc2 Re8 13. Nf1 Bf8 
14. Ng3 g6 15. a4 c5 16. d5 c4 17. Bg5 h6 18. Be3 Nc5 19. Qd2 h5 20. Bg5 Bg7 
21. Nh2 Qc7 22. Rf1 Nh7 23. Bh6 Bh8 24. f4 exf4 25. Bxf4 Nf6 26. Rae1 bxa4 
27. Nf3 Nfd7 28. Bh6 Ne5 29. Nxe5 Bxe5 30. Rf3 Qb6 31. Kh1 Qxb2 32. Ref1 Re7 
33. Bg5 Rd7 34. Bf6 Bxf6 35. Rxf6 a3 36. Nxh5 a2 37. Qh6 gxh5 38. R6f3 h4 
39. Rf4 f5 40. Rxf5 Rg7 41. Rh5 1-0
\end{Verbatim}

\subsection{Forsyth–Edwards Notation (FEN)}
FEN is a notation system used to describe the state of a chess game. It represents the positions of pieces on the chessboard, active color, castling rights, en passant targets, and the half-move and full-move counters. Here is an example of a FEN notation representing the starting position:

\small
\begin{Verbatim}[frame=single]
rnbqkbnr/pppppppp/8/8/8/8/PPPPPPPP/RNBQKBNR w KQkq - 0 1
\end{Verbatim}

In this FEN notation, each letter represents a piece on the board, with uppercase letters representing white pieces and lowercase letters representing black pieces. The forward slash ("/") separates ranks, and the number after each rank indicates the number of empty squares. The active color is represented by "w" for white or "b" for black. The castling rights are denoted by "K" (white kingside), "Q" (white queenside), "k" (black kingside), and "q" (black queenside). The en passant target square is indicated with the corresponding square, or "-" if there is no en passant target. The half-move and full-move counters specify the number of half-moves since the last pawn move or capture and the current move number, respectively.

These different chess formats serve various purposes, from representing individual moves (UCI) to recording entire games (PGN) and describing specific positions (FEN). Understanding and working with these formats is essential for tasks like parsing, analyzing, and exchanging chess game data in different contexts.
\section{Image sources}
\label{apx:image_src}

Here we provide the sources for images used in our~\Cref{fig:main}.
\begin{itemize}[leftmargin=*]
    \item Online game: https://lichess.org/tv
    \item Pro game: https://www.insidethegames.biz/articles/1110959/magnus-carlsen-chess-world-cup-duda
    \item Computer game: https://www.chess.com/news/view/updated-alphazero-crushes-stockfish-in-new-1-000-game-match
    \item Modeling: https://www.chess.com/terms/fen-chess
    \item Puzzle: https://chesspuzzle.net/
    \item General text: https://www.bbc.co.uk/newsround/66233770
    \item Wiki: https://en.wikipedia.org/wiki/Chess
    \item Forum: https://www.chess.com/forum
    \item Blog: https://www.chess.com/blogs
    \item Book: https://www.amazon.co.uk/Chess-Strategy-Action-John-Watson/dp/1901983692
    \item Annotated game: https://www.chess.com/forum/view/game-analysis/kingside-pawn-rush
    \item Youtube video: https://www.youtube.com/watch?v=NAIQyoPcjNM
    \item Reddit: https://www.reddit.com/r/chess/comments/15s65hb/white\_to\_play\_and\_reach\_2000\_chesscom/jwdqpze/?context=3
    \item Value judgement: https://www.thenewatlantis.com/wp-content/uploads/legacy/20190820\_TNA58Wilkenfeldbanner.jpg
\end{itemize}
\section{Dataset analysis}
\label{apx:data_stats}

\begin{table}[t]
\caption{Dataset statistics}
\centering
\begin{tabular}{cccc}
\toprule
Component & Raw size & Document count  & Subset type \\ 
\midrule
Lichess & 19.9 GB & 17.5 M  & Game\\
Pro-player & 0.37 GB & 0.44 M & Game\\
CCRL & 3.60 GB& 3.00 M & Game\\
Chess puzzles & 1.53 GB & 3.19 M& Game\\
Chess modeling & 0.94 GB & 1.90 M & Game\\
\midrule
C4-Chess  & 0.59 GB & 0.18 M& Language\\
Pile-Chess  & 1.10 GB & 0.10 M& Language\\
RedPajama-Chess  & 5.65 GB & 0.52 M & Language \\
Oscar-Chess & 3.13 GB & 0.33 M & Language\\
WikiPedia-Chess & 40.3 MB & 11.4 K& Language\\
Chess Blogs & 0.59 GB & 73.2 K & Language\\
Chess Books & 1.86 GB & 8.36 K & Language\\
Chess Forums & 1.05 GB & 0.14 M & Language\\
\midrule
Annotated Chess Games & 0.66 GB & 245 K & Mixed\\
Youtube transcripts & 0.98 GB & 83.0K & Mixed \\
\midrule
GPT-4-Chess & 0.95 MB & 3.91 K & Conversation\\
Reddit & 0.86 GB & 0.41 M &  Conversation \\
\midrule
Overall & 42.8 GB & 28.1 M& N/A \\
\bottomrule
\end{tabular}
\label{tab:dataset_stats}
\end{table}
\subsection{Dataset statistics and metrics}

\begin{table}[!ht]
    \centering
    \small
    \caption{Metrics for the language dataset.}
    \begin{tabular}{lrrrrrr}
    \toprule
        \multirow{2}*{dataset} & num. words & char. rep. & word rep. & special char & stopwords & perplexity \\
        \multirow{2}* & & ratio & ratio & ratio & ratio & \\
    \midrule
        Chess puzzles & 49.6618 & 0.0104 & 0.0000 & 0.3246 & 0.4771 & 988.3734 \\
        Oscar-Chess & 1441.4341 & 0.0596 & 0.0499 & 0.2242 & 0.4119 & 665.6499 \\
        Pile-Chess & 2105.2454 & 0.0626 & 0.0205 & 0.2409 & 0.4227 & 497.3883 \\
        RedPajama-Chess & 1581.5825 & 0.0532 & 0.0163 & 0.2273 & 0.4218 & 410.1236 \\
        StackExchange-Chess & 578.3733 & 0.0591 & 0.0816 & 0.2617 & 0.4835 & 520.1257 \\
        Wikipedia-Chess & 463.4980 & 0.0876 & 0.0052 & 0.2604 & 0.3079  & 236.9671 \\
        C4-Chess & 510.6041 & 0.0479 & 0.0082 & 0.2131 & 0.4418 & 548.9471 \\
        \bottomrule
    \end{tabular}
    \label{tab:data-analysis-roots}
\end{table}
In\Cref{tab:dataset_stats}, we present the dataset statistics breakdown for each data subset, including its raw size, document count, and subset type.

\Cref{tab:data-analysis-roots} shows the properties of the chess-specific language dataset that we use for training ChessGPT. For these datasets, we computed the average number of words (num. words) per example, character repetition ratio (char. rep. ratio), word repetition ratio (word. rep. ratio.), special character ratio (special char. ratio), stopwords ratio and perplexity of the first shard for each subset in the language dataset. These metrics are based on some of the criteria employed by Bloom~\citep{bigscience2022roots} in their pre-processing pipelines.

\newpage
\subsection{Dataset examples}
In this subsection, we offer dataset examples for each subset to help the readers to understand its contents more intuitively and clearly.
\begin{rbox}{Lichess}
\begin{verbatim}
[Event "Rated Bullet tournament https://lichess.org/tournament/yc1WW2Ox"]
[Site "https://lichess.org/PpwPOZMq"]
[Date "2017.04.01"]
[Round "-"]
[White "Abbot"]
[Black "Costello"]
[Result "0-1"]
[UTCDate "2017.04.01"]
[UTCTime "11:32:01"]
[WhiteElo "2100"]
[BlackElo "2000"]
[WhiteRatingDiff "-4"]
[BlackRatingDiff "+1"]
[WhiteTitle "FM"]
[ECO "B30"]
[Opening "Sicilian Defense: Old Sicilian"]
[TimeControl "300+0"]
[Termination "Time forfeit"]
1. e4 { [%
2. Nf3 { [%
3. Bc4 { [%
4. c3 { [%
5. Bb3?! { [%
6. Bc2 { [%
7. d4 { [%
8. Qxd3 { [%
9. e5 { [%
10. Bg5?! { [%
11. Nbd2?? { [%
12. Bh4 { [%
13. b3?? { [%
\end{verbatim}
\end{rbox}
\begin{rbox}{Pro-player}
\begin{verbatim}
[Event "URS-ch34"]
[Site "Tbilisi"]
[Date "1966.??.??"]
[Round "9"]
[White "Bronstein, David I"]
[Black "Suetin, Alexey S"]
[Result "1/2-1/2"]
[WhiteElo ""]
[BlackElo ""]
[ECO "B97"]

1.e4 c5 2.Nf3 d6 3.d4 cxd4 4.Nxd4 Nf6 5.Nc3 a6 
6.Bg5 e6 7.f4 Qb6 8.Qd2 Qxb2
9.Rb1 Qa3 10.Bxf6 gxf6 11.Be2 Bg7 12.O-O Nc6 
13.Nxc6 bxc6 14.Rb3 Qc5+ 15.Kh1 f5
16.exf5 exf5 17.Na4 Qd4 18.Qxd4 Bxd4 19.Rd1 Bf2 
20.Rxd6 O-O 21.Nb6 Bxb6 22.Rxb6 Re8
23.Bf1 Be6 24.Kg1 Bxa2 25.Rxa6 Rxa6 26.Bxa6 Bd5 
27.Kf2 Re4 28.g3 Bc4 29.Rxc6 Re2+
30.Kg1 Bxa6 31.Rxa6 Rxc2 32.Ra5 Kg7 33.Rxf5 Kg6
34.Rg5+ Kf6  1/2-1/2

\end{verbatim}

\end{rbox}

\begin{rbox}{CCRL}
\begin{verbatim}
[Event "CCRL 40/15"]
[Site "CCRL"]
[Date "2022.01.08"]
[Round "806.6.381"]
[White "Stockfish 060122 64-bit"]
[Black "Dragon by Komodo 2.6 64-bit"]
[Result "1/2-1/2"]
[ECO "D30"]
[Opening "Queen's gambit declined"]
[PlyCount "115"]
[WhiteElo "3505"]
[BlackElo "3480"]

1. d4 {book} d5 {book} 2. c4 {book} e6 {book} 3. Nf3 {book} Nf6 {book}
4. g3 {book} a6 {book} 5. c5 {book} b6 {book} 6. cxb6 {+0.23/33 28s}
c5 {-0.23/30 40s} 7. Bg2 {+0.24/30 11s} cxd4 {-0.15/29 17s} 8. Nxd4
{+0.08/32 15s} Nbd7 {-0.21/29 18s} 9. Nc3 {+0.29/30 18s} Nxb6
{-0.31/29 21s} 10. O-O {+0.27/30 12s} Bb7 {-0.21/28 20s} 11. Be3
{+0.41/31 14s} Nc4 {-0.12/27 19s} 12. Bg5 {+0.31/30 18s} Be7 {-0.16/29
29s} 13. Qa4+ {+0.24/29 18s} Qd7 {-0.08/30 20s} 14. b3 {+0.07/33 22s}
Nd6 {+0.00/29 17s} 15. Rfc1 {+0.19/31 28s} Rc8 {-0.02/31 15s} 16. e3
{+0.13/33 24s} Nde4 {-0.02/33 16s} 17. Qxd7+ {+0.27/34 21s} Kxd7
{-0.05/34 31s} 18. Nxe4 {+0.25/37 13s} Nxe4 {-0.17/32 22s} 19. Bxe7
{+0.29/32 22s} Kxe7 {-0.24/32 20s} 20. Bxe4 {+0.19/35 36s} dxe4
{-0.12/34 37s} 21. Kf1 {+0.13/38 88s} g5 {-0.03/33 27s} 22. Ne2
{+0.00/37 63s} Kd6 {-0.03/36 21s} 23. Rxc8 {+0.08/40 12s} Rxc8
{-0.08/37 25s} 24. Ke1 {+0.12/39 20s} Bd5 {-0.05/34 40s} 25. Rc1
{+0.07/41 17s} Ra8 {-0.06/32 27s} 26. Kd2 {+0.05/33 16s} a5 {+0.00/35
32s} 27. Nc3 {+0.10/35 18s} Rb8 {-0.05/37 20s} 28. Rh1 {+0.04/39 16s}
Ke5 {+0.00/39 67s} 29. Rc1 {+0.07/43 56s} f5 {+0.00/39 21s} 30. Na4
{+0.00/41 13s} Kd6 {+0.00/40 19s} 31. Rc5 {+0.00/46 34s} Ra8 {+0.00/42
57s} 32. Rc2 {+0.00/49 20s} Rb8 {+0.00/40 13s} 33. Nb2 {+0.12/37 17s}
h5 {+0.00/41 14s} 34. Nc4+ {+0.11/41 29s} Bxc4 {+0.00/42 14s} 35. Rxc4
{+0.00/46 23s} h4 {+0.00/43 21s} 36. Ra4 {+0.00/48 20s} Rb5 {+0.00/43
16s} 37. Kc3 {+0.00/50 29s} Ke5 {+0.00/44 18s} 38. b4 {+0.00/51 24s}
axb4+ {+0.00/41 26s} 39. Rxb4 {+0.00/53 30s} Rc5+ {+0.00/43 14s} 40.
Kb3 {+0.00/53 26s} Rd5 {+0.00/43 20s} 41. a4 {+0.00/50 20s} Kd6
{+0.00/45 31s} 42. Rc4 {+0.00/50 16s} h3 {+0.00/48 25s} 43. Kc2
{+0.00/53 34s} Kd7 {+0.00/47 23s} 44. Kb3 {+0.00/47 17s} Rd3+
{+0.00/49 22s} 45. Kc2 {+0.00/50 19s} Rd5 {+0.00/51 28s} 46. Rc3
{+0.00/52 35s} g4 {+0.00/53 22s} 47. Kc1 {+0.00/57 23s} Ra5 {+0.00/55
25s} 48. Rc4 {+0.00/58 21s} Ra6 {+0.00/57 21s} 49. Kc2 {+0.00/59 59s}
Rd6 {+0.00/57 28s} 50. a5 {+0.00/58 14s} Rd5 {+0.00/53 23s} 51. Ra4
{+0.00/60 27s} Kc7 {+0.00/57 16s} 52. a6 {+0.00/60 17s} Kb8 {+0.00/60
18s} 53. a7+ {+0.00/59 25s} Ka8 {+0.00/65 42s} 54. Ra6 {+0.00/59 20s}
Rc5+ {+0.00/66 16s} 55. Kb2 {+0.00/65 25s} Rd5 {+0.00/66 14s} 56. Kc2
{+0.00/62 17s} Rc5+ {+0.00/66 18s} 57. Kd1 {+0.00/67 17s} Rd5+
{+0.00/63 14s} 58. Kc2 {+0.00/101 20s, Draw by 3-fold repetition}
1/2-1/2 
\end{verbatim}
\end{rbox}
\begin{rbox}{Chess puzzles}
\begin{verbatim}
Try your hand at this chess puzzle. The board's FEN is
1r4k1/4nppp/8/4Pb2/8/1P5P/r1PR4/3R3K w - - 0 27, and you need to
determine the optimal move for the player. This puzzle focuses on
backRankMate,endgame,mate,mateIn2,short, and the solutions are
provided in both SAN format as d2d8,b8d8,d1d8 and UCI format as 27.
Rd8+ Rxd8 28. Rxd8#.   
\end{verbatim}
\end{rbox}
\begin{rbox}{Chess modeling}
\begin{verbatim}
With the FEN board state 3r4/4Rpp1/2N2k1p/8/5B2/5BP1/P4PKP/8 b - - 6
42 and a move in UCI d8d6, what is the corresponding SAN move? The
derived move is Rd6.    
\end{verbatim}
\end{rbox}

\begin{rbox}{C4}
\begin{verbatim}
Nothing improves your chess more than playing long time control
tournament events. The Irish championships are being held in Dublin
next weekend. There are a lot of events to choose from! We have a
message encouraging us to play from the Irish Chess Union Chairman.
All members of our club are registered (by the club) with the ICU. You
are eligible to play and should consider playing!
\end{verbatim}
\end{rbox}

\begin{rbox}{Pile}
\begin{verbatim}
Q: What is the theory behind center control? Center control is an
important aspect of playing chess, most openings are built around
controlling the center, but why? Is center control really that
important for winning a game?
\end{verbatim}
\end{rbox}

\begin{rbox}{RedPajama}
\begin{verbatim}
... Mark Dvoretsky - strong player and fantastic coach Mark
Israilewitsch Dvoretsky was born on 9th December 1947 in Moscow. After
finishing his studies of Mathematics and Economics in 1972 Dvoretsky
focused on a career as chess trainer and among other things worked for
Botvinnik's school of chess. As a young player Dvoretsky achieved a
number of notable sucesses: in 1973 he won the Championship of Moscow
and in 1974 he finished fifth at the USSR-Championship in Leningrad.
One year later, in 1975, he won the B-tournament in Wijk aan Zee. But
he soon decided to focus on his career as a chess trainer Dvoretsky
has trained countless strong players, and among his regular students
are well-known players such as Valery Chechov, Nana Alexandria, Sergei
Dolmatov, Alexej Dreev and Artur Jussupow. Among the players who
occasionally trained with Dvoretsky are Garry Kasparov, Viswanthan
Anand, Veselin Topalov, Evgeny Bareev, Viktor Bologan, Loek van Wely
and lots of others. One training method of Dvoretsky was to play
selected positions with both colors against his students - and he
often surprised his students by winning the same position with Black
and with White. Dvoretsky was an International Master and FIDE Senior
Trainer. He published a number of textbooks, sometimes with Artur
Jussupow as co-author. ChessBase published a digital version of his
"Endgame Manual". Dvoretsky was a firm part of chess life in Moscow
and popular guest at chess events all over the world. Russian Chess
Federation ...
\end{verbatim}
\end{rbox}

\begin{rbox}{Oscar}
\begin{verbatim}
If one wishes to learn chess from some of the greatest players in the
world, but does not live in greater New York, then online lessons may
be what he or she is looking for. Many of our coaches are experienced
in teaching both group and solo online lessons. Online lessons are
orchestrated via Skype using an online chess program. 
\end{verbatim}
\end{rbox}
\begin{rbox}{Chess Blogs}
\begin{verbatim}
What Is The Elo Rating System? The Elo rating system measures the relative
strength of a player in some games, such as chess, compared to other players.
Its creator, Arpad Elo, was a physics professor in the United States and a chess
master who worked to improve the way the U.S. Chess Federation measured their 
players' skill levels. He was a solid chess player himself, as you can see from 
this game he played against a young Bobby Fischer.

[Event "New Western Open"]
[Site "Milwaukee, WI USA"]
[Date "1957.07.04"]
[Round "2"]
[White "Arpad Elo"]
[Black "Robert James Fischer"]
[Result "0-1"]
[EventDate "?"]
[ECO "B93"]
[WhiteElo "?"]
[BlackElo "?"]
[PlyCount "98"]

1. e4 c5 2. Nf3 d6 3. d4 cxd4 4. Nxd4 Nf6 5. Nc3 a6 6. f4 e5 7. Nf3 Qc7 8. Bd3
Nbd7 9. O-O b5 10. Qe1 Bb7 11. a3 g6 12. Qh4 Bg7 13. g4 exf4 14. Bxf4 O-O 15.
Qg3 Ne5 16. Nxe5 dxe5 17. Bxe5 Qc5+ 18. Rf2 Nh5 19. Bd6 Qxc3 20. bxc3 Nxg3 21.
Bxf8 Rxf8 22. hxg3 Bxc3 23. Rb1 Bd4 24. a4 Bc8 25. axb5 axb5 26. Rxb5 Bxg4 27.
Kg2 Bxf2 28. Kxf2 Be6 29. Rc5 Kg7 30. Kf3 Kf6 31. Kf4 Ra8 32. g4 h6 33. g5+
hxg5+ 34. Rxg5 Rh8 35. Rg2 g5+ 36. Kf3 Rh3+ 37. Rg3 Rxg3+ 38. Kxg3 Ke5 39. c3
Bd7 40. Bc4 f6 41. Bd5 Be8 42. c4 Kd4 43. Kg4 Bg6 44. Kf3 Bh5+ 45. Kf2 Bd1 46.
Kg3 Be2 47. c5 Kxc5 48. Be6 Kd4 49. Bf5 Ke3 0-1

...

\end{verbatim}
\end{rbox}
\begin{rbox}{Chess Books}
\begin{verbatim}
The following illustrative game is apparently complicated, but it is this in 
its motives\nonly.\nIn reality itis the fight against White's e4 pawn, which
dominates. Shoosmith-\nNimzowitsch, Ostend, 1907. 1.d4 Nf6 2.c4 d63.Nf3 Nbd7
4.Nc3e5 5.e4 Be7 6.Bd3\n0-0 7.0-0 exd4! (if 7...Re8, then 8.05 and Black 
will be\ncramped for along time. For example, 7...Re8 8.45 NcS\n9.Be3 Nxd3
10.Qxd3 Nd7 11.b4 a5 12.43, etc) 8.Nxd4\nRe8 9.b3 Ne5 10.Bc2 a6 (this 
advance will soon be\nintelligible) 11.Bb2 Bd7 12.3 Bf8 13.f4 Ng6 14.Qf3 
c6\n15.Rae1 b5 (now the situation is clear: Black keeps an\neye on White's e-
pawn and seeks ...
\end{verbatim}
\end{rbox}
\begin{rbox}{Chess Forums (StackExchange)}
\begin{verbatim}
Q: Is there a way to use handicaps in chess to bridge the gap between
players of different skill levels? Handicapping is routine in the
Japanese game Go (my best game). The basic strength unit is one stone,
and a one-stone difference represents a full level of difference in
strength. I (about a 1500 player) once asked a 2100 player how much of
a handicap she would need to give me so that we would have an equal
chance to win. "Probably a knight, maybe more," she answered. I once
took a knight handicap against a 2200 player and lost, but it was a
much tighter, closer game than one with no handicap. That might
suggest that a pawn is equivalent to about 200 points of rating.
Apparently handicapping doesn't do much for say, a 50 point difference
in strength (you just play and take your chances). But above that,
there might be ways to use handicaps. Even giving someone the first
move two times out of three (as was earlier done in professional Go)
might do something. Or would it? Why hasn't handicapping been done
much formally in chess, as in Go?
\end{verbatim}
\end{rbox}

\begin{rbox}{Annotated pgn}
\begin{verbatim}
[Event "All about the Traxler Counter-Attack: Why to play Traxler instead of a
passive move"]
[Site "https://lichess.org/study/WLyfoXTJ/xdA6LWme"]
[Date "????.??.??"]
[Round "?"]
[White "?"]
[Black "?"]
[Result ""]
[Annotator "https://lichess.org/@/Dat_Das"]
[ECO "C57"]
[Opening "Italian Game: Two Knights Defense, Fried Liver Attack"]
[UTCDate "2017.10.13"]
[UTCTime "16:33:43"]
[Variant "Standard"]

{ Hello everyone, please click the little heart to show that this
study was helpful to you, to spread the word and to show your
appreciation. } 1. e4 e5 2. Nf3 Nc6 3. Bc4 Nf6 4. Ng5 { You may wonder
why you should play Bc5 instead of d5. This is just to show you
exactly what white is trying to do. } 4... d5 5. exd5 { It seems
you'll win the exchange. } 5... Nxd5 ( 5... Na5 { This is the best
defense if you do play d5. } 6. Bb5+ c6 7. dxc6 bxc6 8. Qf3 ) 6. Nxf7
{ White has a sacrifice of their own. This is the fried liver attack.
} 6... Kxf7 7. Qf3+ Ke6 8. Nc3 { The knight is pinned. } 8... Nb4 9.
a3 Nxc2+ { Sacrifcing a rook. } 10. Kd1 Nxa1 11. Nxd5 { A move with
potential for a dangerous discovered attack } 11... Qh4 12. Nxc7+ {
Double check. } 12... Kd7 13. Qf7+ Qe7 14. Nxa8 Qxf7 15. Bxf7 { And
Black's king position is destroyed, and white is a pawn up. White's
knight may be hanging, but so is black's. }

\end{verbatim}
\end{rbox}

\begin{rbox}{Youtube transcripts dataset}
\begin{verbatim}
{'text': 'sink field cup round number six coverage', 'start': 
0.96,'duration': 4.319}, {'text': "i'm going to recap all four games for", 
'start': 3.6, 'duration': 4.08}, {'text': 'you guys not a lot of drama today 
on the', 'start': 5.279, 'duration': 4.961}, {'text': "hans niemann front so 
let's just jump", 'start': 7.68, 'duration': 4.56}, {'text': 'right into the
games three decisive', 'start': 10.24, 'duration': 3.92}, {'text': 'games the
first game to finish was janna', 'start': 12.24, 'duration': 3.68}, {'text':
'pomniacci with the white pieces against', 'start': 14.16, 'duration': 3.6},
{'text': 'fabiano carawana', 'start': 15.92, 'duration': 4.72}, {'text': 'and
we see a sicilian by fabi', 'start': 17.76, 'duration': 5.359}, {'text': 'and 
he plays the tymanov variation with', 'start': 20.64, 'duration': 4.96}, 
{'text': 'e6 sometimes this is called the sicilian', 'start': 23.119, 
'duration': 4.721}, {'text': "con it's a very solid sicilian i've", 'start': 
25.6, 'duration': 3.919}, {'text': 'played this quite a bit myself from the',
'start': 27.84, 'duration': 3.279}, {'text': 'black side', 'start': 29.519, 
'duration': 2.961}, {'text': 'takes takes', 'start': 31.119, 'duration': 
3.041}, {'text': 'knight f6', 'start': 32.48, 'duration': 3.919}, {'text': 
'and now we get kind of this uh', 'start': 34.16, 'duration': 4.559}, 
{'text': 'taimanov set up with the knight on f6', 'start': 36.399, 
'duration': 4.241}, {'text': 'instead of queen c7', 'start': 38.719, 
'duration': 3.68}, {'text': 'so this can sometimes transpose into', 'start': 
40.64, 'duration': 3.2}
...
\end{verbatim}
\end{rbox}

\begin{rbox}{GPT4 instruction-tuning data}
\begin{verbatim}
{
  "conversations": [
    {
      "author": "Human 0", 
      "text": "Explain the difference between 
        the Italian Game and the Ruy Lopez.", 
      "url": "gpt-4", 
      "score": 0
    }, {
      "author": "Human 1",
      "text": "The Italian Game starts with 1.e4 e5 2.Nf3 Nc6
          3.Bc4, focusing on targeting the f7 square. 
          The Ruy Lopez begins with 1.e4 e5
          2.Nf3 Nc6 3.Bb5, attacking the knight on c6 to 
          undermine Black's pawn structure
          and exert pressure on the e5 pawn. 
          The Italian Game tends to be more tactical,
          while the Ruy Lopez is more strategic and positional.",
      "url": "gpt-4",
      "score": 0
    }
  ]
}
\end{verbatim}
\end{rbox}

\begin{rbox}{Reddit Conversation data}
\begin{verbatim}
{
  "author": "Human 0",
  "text": "Honest question for those with higher ELOs. First, sorry if
  this is a confusing or vague question but I'll try my best to word
  it: At what rating, in your opinions, do you find people stop making
  \"silly\" mistakes? I'm at 1500 rapid and 1300 blitz, trying to
  improve my play. I found that at these ratings blunders are rare
  (assuming adequate time). It seems like mostly people just don't
  see/miss tactics or play inaccuracies rather than playing straight
  up mistakes/blunders. Do you higher rated elo players feel the same
  way? Or do you think the inaccuracies we make/tactics we miss are
  quite obvious, the same way I can see a blunder is obvious? Curious
  on the perspective.",
  "score": 8,
  "other_data": false,
  "url": "/r/chess/comments/nhh2mn/honest_question_for_those_with_higher_elos/"
}
{
  "author": "Human 1",
  "text": "Im 2200 in blitz on lichess, and of my last 10 blitz games,
  around 8 were decided by major blunders. In rapid, the amount of
  major blunders decreases a lot, but they are still very common. When
  you get higher rated, you will still blunder because you are also
  going to be facing higher rated opponents. If I played a 1200, I
  would rarely blunder but I blunder very easily against 2300+
  people.",
  "score": 22,
  "other_data": false,
  "url": "/r/chess/comments/nhh2mn/honest_question_for_those_with_higher_elos/
  gywapvb/"
}
\end{verbatim}
\end{rbox}

\section{Data Collection and Preprocessing}
\label{apx:data}
\subsection{Online Chess websites}
\label{apx:online_chess_web}
We choose around 26 chess websites, including chess.com and Lichess.com to scrape chess-related language corpus. We gather a diverse range of chess and language data from various platforms, including blogs, news, and articles. Specifically, we focused on extracting relevant information from several topics including blogs, news, openings, chess terms, forums, and articles. These topics were carefully chosen as they contain valuable texts pertaining to chess background knowledge, news, and also PGN games in some instances. 

We utilize Beautifulsoup\footnote{\url{https://www.crummy.com/software/BeautifulSoup/}}and Playwright \cite{playwright} to parse HTML pages and locate all the texts and PGNs. We further transfer those PGNs into text which helps us build mixed game-language datasets from these sources. We record crucial details such as the URL, author, title, time, and the actual content of the articles. 

\subsection{Online Chess forums}
We choose 5 chess forums and follow basically similar way with~\Cref{apx:online_chess_web} to scrape forum text data.

\subsection{Annotated PGN}
We collect our annotated PGN from 7 sources: Lichess studies, Chess publishing, Megabase, Pgnlib, pathtochessmastery and gameknot.

\textbf{Lichess studies} Lichess Study provides a rich collection of annotated PGNs. The annotations are embedded in PGNs, explaining the insight of the moves. Users can conveniently search for studies based on keywords or specific topics like Sicilian Defense, Puzzles, or d5. To enhance the searching process, we collect a comprehensive set of 54 popular keywords. Our implementation leverages Selenium's \cite{selenium} built-in functions to efficiently parse HTML pages and simulate the searching process. Additionally, we use Lichess APIs\footnote{\url{https://lichess.org/api}}to request for PGN games associated with a specific study ID.

\textbf{Chess publishing} This contains commercial annotated PGNs from Chesspublishing.com\footnote{\url{https://www.chesspublishing.com/content/}}, so we do not open source this source of data.

\textbf{Megabase} This contains commercial annotated PGNs from Megabase2023\footnote{\url{https://shop.chessbase.com/en/products/mega_database_2023}}, so we do not open source this source of data.

\textbf{Pgnlib} We collect annotated PGN from Pgnlib\footnote{\url{https://www.angelfire.com/games3/smartbridge/}}.

\textbf{Pathtochessmastery} We collect annotated PGNs from Path to Chess Mastery\footnote{\url{https://www.pathtochessmastery.com/}}.

\textbf{Gameknot} We use Selenium \cite{selenium} to scrape annotated PGNs from gameknot\footnote{\url{https://gameknot.com/list_annotated.pl?u=all}}.

\subsection{Existing Datasets}
\label{apx:existing_datasets}
We mainly extract all chess-related language corpus from existing dataset C4~\citep{2019t5}, Pile ~\citep{gao2020pile}, Oscar~\citep{OrtizSuarezSagotRomary2019}, Wikipedia~\cite{wikidump} and RedPajama~\cite{together2023redpajama}. To extract chess-related language corpus, we first filter language corpus that contains the keyword 'chess'. We further utilize the deberta-v3-base-tasksource-nli model\footnote{\url{https://huggingface.co/sileod/deberta-v3-base-tasksource-nli}}, which can conduct zero-shot classification based on classes specified by users. We set two classes: chess-related and non-chess-related for this model. We input the first 2000 characters of each text example to the model and set the threshold as 0.5.
\subsection{YouTube transcripts dataset}
To collect the youtube transcripts dataset, we first gather 19 keywords and utilize scrapetube\footnote{\url{https://github.com/dermasmid/scrapetube}} to gather all related channels. We extract all videos from these channels and use the same deberta-v3-base-tasksource-nli model mentioned in~\Cref{apx:existing_datasets} to filter all video transcripts and also the corresponding videos that are not relevant to chess. It is fairly easy to extract transcripts from videos and the main difficulty is to extract the FEN chessboard representations from videos. Here we mainly utilize two steps to extract specific FEN from chess videos.
\subsubsection{Extract Chess Board from videos}
The first step is to extract the chessboard from videos. We utilize GLIP~\citep{li2021grounded}, a zero-shot language-based detection model. By setting the prompt as 'Chess, chessboard', we can extract chess-related detection bounding boxes from videos. We conduct further filterings such as length-width ratio filtering to guarantee it is a valid chessboard bounding box in most cases, which will be processed based on the second step.
\subsubsection{Convert Chess board image to FEN text format}
Our second step involves converting the chessboard image into FEN text. FEN format serves as a great way to describe the current chess board state in text-based format. The pipeline that converts the chess board image to FEN format includes three main procedures - chess board decomposition, piece classification, and the prediction of which player is the next turn.

\textbf{Chess board decomposition} The aim of this section is to breakdown a whole chess board into 64 small tiles (8 rows * 8 columns), where each tile contains only one chess piece. To achieve this, we initially convert the RGB image to grayscale, preparing for the line detection process. Subsequently, we make two convolutional kernels to find horizontal and vertical gradients\footnote{\url{https://github.com/Elucidation/tensorflow_chessbot/blob/master/tensorflow_compvision.ipynb}}. The Hough Transform is then applied to detect vertical and horizontal lines and filter out seven vertical and seven horizontal lines that fit the demand. Finally, we divide the board into 64 tiles by having the position of the 14 lines.

\textbf{Piece classification} To facilitate model training and evaluation, we employ an open source chess image dataset on Kaggle\footnote{\url{https://www.kaggle.com/datasets/koryakinp/chess-positions}} which contains 80k training images and 20k testing images.  Each tile can be classified into one of the 13 categories (p, r, b, n, k, q, P, R, B, N, K, Q, and space) which is detailed in Appendix B.3. We implement a model in PyTorch which uses a pre-trained ResNet18~\citep{he2016resnet} due to its well-established performance in image classification tasks. To adapt the model to our specific problem, we replaced the original fully connected layer with a new layer consisting of 13 output neurons, corresponding to the 13 pieces categories. We train the model on 40000 images with hyperparameters shown in \Cref{tab:hyperparameters}.

\begin{table}[t]
    \caption{Training hyperparameters for chess board to FEN}
    \centering
    \label{tab:hyperparameters}
    \small
    \begin{tabular}{lr}
      \toprule
      Hyperparameter   & Value \\
      \midrule
      Batch size &  {32} \\
      Number of epochs & {10} \\
      Learning rate & {0.001} \\
      Optimizer & {SGD} \\
      Momentum & {0.9} \\
      \bottomrule
    \end{tabular}
\end{table}

After the training process, we evaluate the model on a testset with 20k images (equivalent to 128k tiles). Please refer to \Cref{tab:piece_accuracy} for the final accuracy of each category.

\begin{table}[H]
    \caption{Validation accuracy of each chess piece}
    \centering
    \label{tab:piece_accuracy}
    \small
    \begin{tabular}{llr}
      \toprule
      Color & Piece   & Accuracy ($\%$) \\
      \midrule
      Black & \textbf{p}awn &  $99.98$ \\
       & \textbf{r}ook & $99.99$ \\
       & k\textbf{n}ight & $99.98$ \\
       & \textbf{b}ishop & $99.98$ \\
       & \textbf{q}ueen & $100.00$ \\
       & \textbf{k}ing & $99.97$ \\
       \midrule
       White & \textbf{P}awn & $100.00$ \\
       & \textbf{R}ook & $99.98$ \\
       & k\textbf{N}ight & $99.98$ \\
       & \textbf{B}ishop & $99.99$ \\
       & \textbf{Q}ueen & $99.95$ \\
       & \textbf{K}ing & $99.98$ \\
       \midrule
       & Space & $100.00$ \\
      \bottomrule
    \end{tabular}
\end{table}

\textbf{Prediction of next turn} As FEN format also includes the prediction of the next turn which is indicated by “w” for white, and “b” for black, the prediction of the next turn is accomplished by analyzing the main color of each tile. We use Colorthief, a library for grabbing the color palette from images, to extract the main color from each tile since the background color of a tile will be highlighted if a move is played on that tile. Hence, we find the highlighted tile by analyzing the tile color to know who is the current player and naturally infer who is the next turn.

Finally, we also provide a final certainty percentile to evaluate to what extent the generated FEN is correct by calculating the product of the accuracy of the 64 tiles.

\subsection{Lichess dataset}
We collect 5 months of Lichess dataset from the Lichess database \cite{lichess}: 2013-02, 2014-02, 2015-02, 2016-02, and 2017-02. In fact, there are much more data available and we leave more game data for future work. 
\subsection{Pro-player dataset}
We collect our pro-player dataset from PGN Mentor\footnote{\url{https://www.pgnmentor.com/}}.
\subsection{Chess books}
We select 100 chess-related keywords and search for all related chess books (around 9k books) on the online PDF library. Because of the legal issues about books' copyright, we choose not to open-source this source of data. Instead, we only open source the list of books we use.
\subsection{CCRL}
We collect our CCRL dataset without comments from the official website\footnote{\url{https://ccrl.chessdom.com/ccrl/4040/}} for three settings of time control: CCRL BLITZ, CCRL 40/2 FRC and CCRL 40/15.
\subsection{Chess puzzles}
We collect our chess puzzles from the Lichess puzzle dataset\footnote{\url{https://database.lichess.org/\#puzzles}}.
\subsection{Chess modeling data}
We design 11 modeling tasks to generate data:
\begin{itemize}
    \item Given PGN, generate FEN representation.
    \item Given a list of UCI moves, generate FEN representation.
    \item Given FEN and a UCI move, transfer the move to SAN format.
     \item Given FEN and a SAN move, transfer the move to UCI format.
     \item Given FEN, generate an ASCII board.
     \item Given FEN and a UCI move, generate the next FEN.
     \item Given FEN and a SAN move, generate the next FEN.
    \item Given FEN, generate all legal moves in SAN format.
     \item Given FEN, generate all legal moves in UCI format.
     \item Given PGN, generate all legal moves in SAN format.
    \item Given PGN, generate all legal moves in UCI format.
\end{itemize}
To generate the synthetic modeling dataset for these tasks, we utilize PGN data extracted from the first 1 million games of the Lichess database dump from March 2017. In order to encompass a wider range of ELO ratings, we divide the elo-rating into 9 intervals: 0-1000, 1000-1200, 1200-1400, 1400-1600, 1600-1800, 1800-2000, 2000-2200, 2200-2400, and 2400-3000. Random sampling is employed to select games from each interval, ensuring that our dataset contains approximately 10,000 games for each ELO interval. Consequently, the dataset achieves a balanced representation across different ELO ratings. Then we further utilize the python-chess library \cite{python_chess} to complete all the tasks we design to generate our final synthetic modeling dataset.
\subsection{Preprocessing}
We preprocess the data sources in three levels of granularity. 
For sources where existing \textbf{preprocessed dataset} are available, we filter out the subset that contains chess-related information without performing additional preprocessing. 
For sources that we retrieve from the \textbf{Internet}, we only parse portions of the HTML that contains information about chess. We implement different parsers for the different sources we consider. As a result, our data preprocessing can be more light-weight compared to previous work that extracts corpora from raw HTML web pages. 
For \textbf{PGN games}, we use the original PGN but filter out some annotations that are not useful for learning the model. 

The different sources contain data in different formats. To facilitate training on all datasets, we preprocess all datasets to have the same jsonl format. 

We implement the data-preprocessing step for each source as a separate Apache Beam pipeline which allows us to process the datasets in parallel into multiple shards. We note that a simple Apache Beam pipeline implementation provides no guarantees that data processing will be in the same order as they were read from the files. As a result, running the same pipeline twice will produce a set of shards that are shuffled differently. To provide determinism in our data-processing pipeline, we adopt a deterministic shuffling strategy similar to the implementation in TensorFlow Datasets (TFDS) to ensure reproducible data processing while maintaining scalability.

We provide further details on the preprocessing used for each individual source below:

\textbf{C4, Oscar, The Pile, RedPajama, Wikipedia} Since these datasets 
are available in processed format, we do not perform any additional preprocessing.

\textbf{StackExchange} We use the forums Chess StackExchange\footnote{\url{https://chess.stackexchange.com/}}.
We preprocess the StackExchange data in the same way as done in RedPajama. Concretely, for each question, we first clean up the text via simple regular expressions and remove some HTML tags in the text. Then we prefix the question with \texttt{Q:} and the answers with \texttt{A:} and then concatenate the questions with all answers.

\textbf{Chess puzzle and Chess modeling data} The original data format is in CSV format with key data, such as puzzle FEN and puzzle answer. We leverage some language templates to transfer the CSV as natural language text.

\textbf{Lichess database, CCRL and pro-player dataset} We keep only games with known finish, i.e., (win, lose or draw). We remove \texttt{clk}, \texttt{arrow}, \texttt{evp} annotations from the comments. We further remove Emojis from the comments. Afterward, each PGN is considered a single string of text that is used for downstream training.

\textbf{Medium article, Chess forum, Chess books} We run the same preprocessing pipeline as in ~\citep{bigscience2022roots}.

\textbf{Annotated PGNs} We conduct language filtering using Fasttext\footnote{\url{https://fasttext.cc/docs/en/language-identification.html}} in ChessCLIP preprocessing. And we conduct the same preprocessing as we do in Lichess database in ChessGPT training preprocessing.

\textbf{Insturction data from GPT-4} No-further pre-processing.

\textbf{Conversational data from Reddit} We filter the conversations based on the language, response length, number of emojis, blacklist words, and scores.

We initially applied the same data-processing procedure described in
~\citep{bigscience2022roots} for all of the data that we collected. However, we found that the filtering used in~\citep{bigscience2022roots} can be too aggressive in removing useful examples as many of our data sources include a significant portion of chess notation that does not resemble natural language (e.g., chess puzzles). Therefore, we opted for more light-weight pre-processing and use the processing from~\citep{bigscience2022roots} only in cases where the text includes a significant portion of natural language description (blogs for example). In addition, for further protection of privacy, we anonymize user names and replace them with terms like 'Human 0' in all conversation-like data, especially in chess forums and Reddit conversational data.
\subsection{Licenses and dataset cards}
For specific Licenses and dataset cards, refer to our open-source dataset repository: \url{https://huggingface.co/datasets/Waterhorse/chess_data}.

\section{Implementation and Evaluation Details}
We open source all our models: ChessCLIP (\url{https://huggingface.co/Waterhorse/ChessCLIP}), ChessGPT-Base (\url{https://huggingface.co/Waterhorse/chessgpt-base-v1}) and ChessGPT-Chat (\url{https://huggingface.co/Waterhorse/chessgpt-chat-v1}). Refer to these URLs for model licenses and model cards.
\label{examples}
\subsection{Implmenetation details}
\subsubsection{ChessCLIP}
\label{apx:chessclip_details}
For the ChessCLIP dataset, we preprocess the annotated PGNs to produce board/text pairs which we feed separately to the board and text encoders. In particular, for every move in the PGN, we extract the comments attached to the move as well as the board state. While our YouTube transcripts dataset can also serve as training data for ChessCLIP, we have discovered that it consistently contains more noise compared to the annotated PGN dataset. To ensure the stability of our training process, we have chosen to exclusively utilize the annotated PGN datasets. The task of refining the YouTube transcripts for future training remains a part of our ongoing work.

For the ChessCLIP model, we instantiate a ChessCLIP model with a pair of text encoder and a board/action encoder. For board/action encoder, we use a ResNet~\citep{he2016resnet} architecture that conditions the action encoding via a modified FiLM layer~\citep{perez2018film}. We encode the board positions and moves using the same scheme as those used by Leela Chess Zero (lc0)~\citep{LeelaChessZero}, which is similar to the encoding used by AlphaZero~\citep{silver2018az} for encoding positions and moves in chess. Concretely, the board positions are encoded as a $\mathcal{R}^{8 \times 8 \times 112}$ feature map and the actions are encoded as a $\mathcal{R}^{1858}$ vector. For the text encoder, we follow the same architecture as with the original OpenAI CLIP model and we only fine-tune the last two layers of pretrained OpenAI text encoder. Our implementation is based on the open-source implementation \footnote{\url{https://github.com/mlfoundations/open_clip}} of CLIP. We show our training hyper-parameters in~\Cref{tab:ChessCLIP_hyper}.

\begin{table}[H]
    \centering
    \small
    \caption{ChessCLIP Training Hyperparameters}
    \label{tab:ChessCLIP_hyper}
    \begin{tabular}{lrlrlr}
      \toprule
      Hyperparameters    & Value & Hyperparameters    & Value & Hyperparameters    & Value\\
      \midrule
      Learning Rate &  {5e-4} & Warmup Step & {500} & Weight decay & {0.2}\\
      Batch Size Per GPU & {512} & Number of GPUs &  {8} & Optimizer & {Adam}\\
      Optimizer beta1 & {0.9} & Optimizer beta2 &  {0.98} & Optimizer epsilon & {1e-6} \\
      Precision & {AMP} & Learning Rate Scheduler  &  {Cosine} & Epochs & 40\\
      \bottomrule
    \end{tabular}
\end{table}
We would like to highlight that ChessCLIP can serve as a direct move sequence generator when provided with a text prompt. By utilizing beam search over all legal sequences, we can maximize the similarity between sequences. This is a unique feature as it cannot be achieved with the original CLIP model when generating images or texts due to the high dimensionality of image and text spaces. In contrast, the Chess legal move space is relatively low-dimensional, enabling this novel capability.
\subsubsection{ChessGPT}
\label{apx:chessgpt_details}
We follow common implementations of training a domain-specific instruction-following LLM. Firstly we conduct base-model fine-tuning using chess corpus introduced in section~\ref{game_dataset},~\ref{language_dataset} and ~\ref{mixed_dataset}. Due to computational constraints, we choose to finetune the RedPajama-3B-base~\cite{together2023redpajama} model, which is an open-souce replication of LLaMA~\cite{touvron2023llama}. We also limit our model max token length as 1024. The base-finetuning brings us our base model: \textbf{ChessGPT-Base}. 

After base-finetuning, we conduct supervised fine-tuning by supervised learning on question/conversation response using data introduced in section~\ref{conversation_data} and general conversation data from OASST1~\citep{kopf2023openassistant}, Dolly2 ~\citep{dolly2}, Alpaca-GPT4~\citep{peng2023instruction}, and Sharegpt ~\citep{sharegpt}, forming our chat model: \textbf{ChessGPT-Chat}. We call it \textbf{ChessGPT-Chat} instead of \textbf{ChessGPT-SFT} because some of our conversation datasets are generated by RLHF-tuned LLM. We convert all our Q/A or conversation data into the following two conversation formats:

\textbf{Between two people}: \textit{A friendly, helpful chat between some humans.<|endoftext|>Human 0: \{Human 0 Question\}<|endoftext|>Human 1: \{Human 1 Response\}<|endoftext|>Human 0: \{Human 0 Question\}<|endoftext|>...}

\textbf{Between multiple people (Reddit conversational data)}: \textit{A friendly, helpful chat between some humans.<|endoftext|>Human 0: \{Human 0 Question\}<|endoftext|>Human 1: \{Human 1 Response\}<|endoftext|>Human 2: \{Human 2 Question\}<|endoftext|>...}

Our base-training code refers to llama-finetune\footnote{\url{https://github.com/chaoyi-wu/Finetune_LLAMA}} and our sft-training code refers to the Alpaca~\citep{alpaca} and Fastchat~\citep{vicuna2023}. The training hyperparameters for ChessGPT-Base and ChessGPT-Chat are shown in~\Cref{tab:Chessgpt_base_hyper} and~\Cref{tab:Chessgpt_chat_hyper}.

\begin{table}[H]
    \centering
    \small
    \caption{ChessGPT-Base Training Hyperparameters}
    \label{tab:Chessgpt_base_hyper}
    \begin{tabular}{lrlrlr}
      \toprule
      Hyperparameters    & Value & Hyperparameters    & Value & Hyperparameters    & Value\\
      \midrule
      Learning Rate &  {8e-5} & Warmup ratio & {0.03} & Weight decay & {0.00}\\
      Batch Size Per GPU & {3} & Number of GPUs &  {8} & Optimizer & {Adam}\\
      Accumulation step & {8} & Max token length &  {1024} & Acceleration & {FSDP} \\
      Precision & {bf16} & Learning Rate Scheduler  &  {Cosine}  & Epochs & 1\\
      \bottomrule
    \end{tabular}
\end{table}

\begin{table}[H]
    \centering
    \small
    \caption{ChessGPT-Chat Training Hyperparameters}
    \label{tab:Chessgpt_chat_hyper}
    \begin{tabular}{lrlrlr}
      \toprule
      Hyperparameters    & Value & Hyperparameters    & Value & Hyperparameters    & Value\\
      \midrule
      Learning Rate &  {2e-5} & Warmup ratio & {0.03} & Weight decay & {0.00}\\
      Batch Size Per GPU & {4} & Number of GPUs &  {8} & Optimizer & {Adam}\\
      Accumulation step & {8} & Max token length &  {1024} & Acceleration & {FSDP} \\
      Precision & {bf16} & Learning Rate Scheduler  &  {Cosine}  & Epochs & 1\\
      \bottomrule
    \end{tabular}
\end{table}

Here we also show more evaluation results.

\textbf{General policy result.}
Table \ref{apx:tab_general_black} presents the results of the general policy experiment using black chess, which align with the findings from the previous white chess experiment. The comparison between the two ChessGPT models across different Elo ratings reveals a lack of noticeable distinctions, indicating the model's limited sensitivity to the key information provided in the prompt. A more intuitive illustration of this observation will be provided in the subsequent paragraph. There are two notable points to highlight. Firstly, ChessGPT demonstrates improvements compared to its base model RedPajama and performs on par with LLAMA. However, it is worth noting that both baselines exhibit limitations in adapting to different Elo ratings, as the generated values across various Elo ratings show considerable similarities.
\begin{table}[H]
\caption{General policy evaluation in Black}
\label{apx:tab_general_black}
\small
\centering
\begin{tabular}{ccccc}
\midrule
& \multicolumn{4}{c}{Move Scores (\%)} \\ 
\cmidrule{2-5}
Elo Rating & LLAMA & RedPajama & ChessGPT-Base & ChessGPT-Chat   \\ 
\midrule
700-1000  &   {52.9 $\pm$ 0.9}  &      {46.2 $\pm$ 1.0}            &   51.9 $\pm$ 0.1  & 52.1 $\pm$ 0.9                                            \\
1200-1500   &    {53.2 $\pm$ 0.9}       &      {46.9 $\pm$ 0.9}          &  53.0 $\pm$ 1.0   & {52.4 $\pm$ 1.0}                                    \\ 
1700-2000   &    {52.1 $\pm$ 0.8}       &      {46.6 $\pm$ 1.0}          &  52.0 $\pm$ 1.0   & {52.0 $\pm$ 1.0}                                      \\ 
2700-3000    &    {53.6 $\pm$ 0.9}       &      {47.3 $\pm$ 1.0}          &  52.2 $\pm$ 0.9   & {52.1 $\pm$ 1.1}                                     \\ 
\bottomrule
\end{tabular}
\vspace{-0.2in}
\end{table}

\begin{figure}[H]
    \centering
    \includegraphics[width=\textwidth]{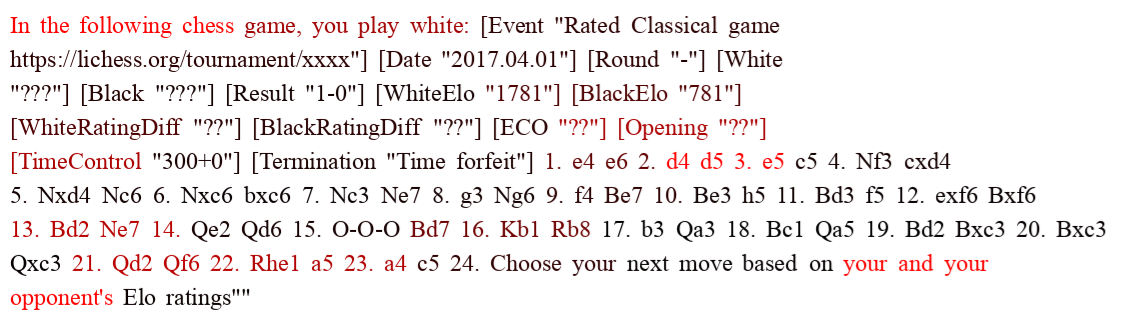}
    \caption{Visualization of ChessGPT-Base Attention: The figure illustrates the attention space of ChessGPT for the General Policy experiment, generating a compound level next move based on \textbf{Elo rating}. The highlighted areas represent the importance of attention, with color intensity ranging from black to red, where red indicates the highest importance."}
    \label{fig:visual}
\end{figure}

\textbf{Words attention visualization.}
To evaluate whether the ChessGPT-Base model captures the key information in the general policy task, we conducted a visualization analysis of its self-attention mechanism. The visualization, as shown in Figure \ref{fig:visual}, reveals that the model does attend to the "WhiteElo" and "BlackElo" values to some extent. However, the level of attention dedicated to these important features appears to be relatively weak. This suggests that the model's ability to appropriately incorporate and utilize the Elo ratings during the generation process is not as strong as desired. Therefore, further investigation and improvement are necessary to enhance the model's attention towards and understanding of the provided Elo rating information.
\subsubsection{Compute resources}
We use 8$\times$80G A100 GPUs for all our experiments. It takes 5 hours to train ChessCLIP using all A100 GPUs. And it takes 60 hours to train ChessGPT-Base model and 18 hours to train ChessGPT-Chat.
\label{apx:compute}
\section{Evaluation details}

\subsection{Evaluation task examples and plots}
Here we show task examples and plots for our evaluation tasks. Basically, the evaluation tasks consist of three parts: Task Prefix, which can be regarded as a description of the task and is also the main prompt we use for LLMs. Input, which is the question and the input of LLMs. And Target, which contains the answer of the question for exact string match tasks, or target score, which provides the score for each available answer for multi-choice tasks.
\begin{figure}[H]
    \centering
    \includegraphics[width=0.5\linewidth]{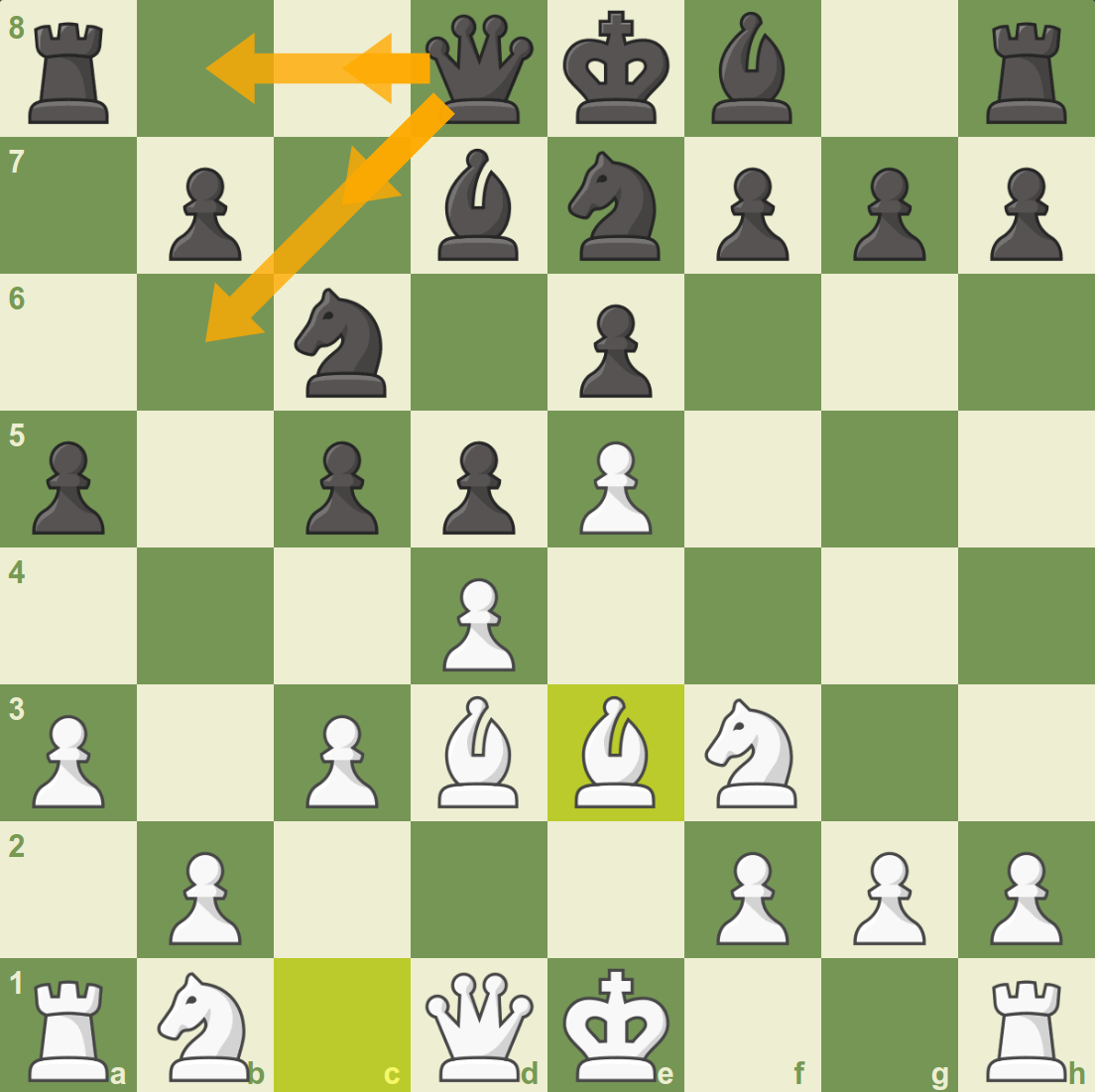}
    \caption{\textbf{Example in Chess state tracking.\protect\footnotemark} The input leads to the following board configuration.The task is to predict the squares to which the piece at d8, the black queen, can be legally moved. Here the black queen at square d8 can be legally moved to any of the squares ["b8", "b6", "c7", "c8"].}
    \label{fig:CST}
\end{figure}
\footnotetext{{\url{https://github.com/google/BIG-bench/tree/main/bigbench/benchmark_tasks/chess_state_tracking}}}
\begin{rbox}{The Chess state tracking (Fig.\ref{fig:CST})}
\begin{itemize}[leftmargin=*]
\item \textbf{Task Prefix}: For each of the following (in-progress) chess games, please complete the notation for the last shown move by filling in the destination square.

\item \textbf{Input}: e2e4 e7e6 d2d4 d7d5 e4e5 c7c5 c2c3 b8c6 g1f3 g8e7 a2a3 a7a5 f1d3 c8d7 c1e3 d8

\item \textbf{Target}: ["b8", "b6", "c7", "c8"]
\end{itemize}
\end{rbox}

\begin{rbox}{Board state tracking}
\begin{itemize}[leftmargin=*]
\item \textbf{Task Prefix}: Could you produce the Forsyth–Edwards Notation (FEN) that corresponds to the provided PGN-based move list of the chess games?

\item \textbf{Input}: 1. d4 g6 2. c4 Bg7 3. e4 Nf6 4. Nc3 O-O 5. Be3 Ne8 6. f3 Nc6 7. Qd2 e6 8. h4 d5 9. cxd5 exd5 10. Nxd5 Nf6 11. Nxf6+ Bxf6 12. e5 Bg7 13. Bb5 Bd7 14. Rc1 a6 15. Bc4 b5 16. Bb3 Bf5 17. Rxc6 Qd7 18. Rc5 c6 19. Ne2 Rad8 20. Ng3 Be6 21. Bxe6 Qxe6 22. Ne4 Rd5 23. O-O f5 24. Rxd5 cxd5 25. Nc5 Qc6 26. Bh6 f4 27. Bxg7 Kxg7 28. Re1 Qe8 29. e6 Qe7 30. Qa5 Qxh4 31. e7 Re8 32. Nxa6 Rxe7 33. Rxe7+ Qxe7 34. Nc7 Qe3+ 35. Kh2 Qf2 36. Nxd5

\item \textbf{Target}: 8/6kp/6p1/Qp1N4/3P1p2/5P2/PP3qPK/8 b - - 0 36
\end{itemize}
\end{rbox}

\begin{rbox}{Board state tracking}
\begin{itemize}[leftmargin=*]
\item \textbf{Task Prefix}: Could you produce the Forsyth–Edwards Notation (FEN) that corresponds to the provided SAN-based move list of the chess games?

\item \textbf{Input}: e2e4 d7d5 b1c3 d5e4 c3e4 g8f6 e4c3 e7e6 f1c4 f8b4 d2d3 b4c3 b2c3 e8g8 g1e2 c7c5 e1g1 b8c6 e2g3 e6e5 a2a4 c8g4 f2f3 g4f5 c1g5 f5g6 d1d2 h7h6 g5h6 g7h6 d2h6 c6e7 f3f4 e5f4 f1f4 d8d6 a1f1 f6h7 g3h5 g6h5 h6d6 h5g6 d6e7 a8e8 e7h4 e8e2 h4h6 e2c2 h6g6 g8h8 c4f7 c2

\item \textbf{Target}: 5r1k/pp3B1n/6Q1/2p5/P4R2/2PP4/2r3PP/5RK1 b - - 0 26
\end{itemize}

\end{rbox}

\begin{rbox}{{Chess state value}}
\begin{itemize}[leftmargin=*]
\item \textbf{Task Prefix}: Evaluate the following PGN to see whether black or white takes advantage.

\item \textbf{Input}: 1. e4 e6 2. d4 d5 3. e5 c5 4. Nf3 cxd4 5. Nxd4 Nc6 6. Nxc6 bxc6 7. Nc3 Ne7 8. g3 Ng6 9. f4 Be7 10. Be3 h5 11. Bd3 f5 12. exf6 Bxf6 13. Bd2 Ne7 14. Qe2 Qd6 15. O-O-O Bd7 16. Kb1 Rb8 17. b3 Qa3 18. Bc1 Qa5 19. Bd2 Bxc3 20. Bxc3 Qxc3 21. Qd2 Qf6 22. Rhe1 a5 23. a4 c5

\item \textbf{Target Score}: \{"Black has advantage.": 1, "The game is equal.": 0, "White has advantage.": 0\}
\end{itemize}
\end{rbox}

\begin{rbox}{Chess annotation}
\begin{itemize}[leftmargin=*]
\item \textbf{Task Prefix}: Annotate the last step of the following PGN.

\item \textbf{Input}: 1. d4 Nf6 2. c4 e6 3. Nf3 Bb4+ 4. Bd2 a5 5. g3 O-O 6. Bg2 b6 7. O-O Ba6 8. Bg5 Be7 9. Qc2 Nc6 10. a3 h6 11. Bxf6 Bxf6 12. Rd1 Qe7 13. e3 Rae8 14. Nfd2 g5

\item \textbf{Target Score}: {"Karpov could have resigned here with a clear conscience.": 0, "White intends to further weaken Black's kingside with 19.h5.": 0, "20...Kh7 21.Bxg6+ fxg6 22.Qxe6 Gives White a winning attack.": 0, "Black overreacts to the positional strength of White's game. 14...g6 would have been more solid.": 1}
\end{itemize}
\end{rbox}

\begin{rbox}{Chess opening (opening2pgn)}
\begin{itemize}[leftmargin=*]
\item \textbf{Task Prefix}: Show me the PGN of the following opening.

\item \textbf{Input}: Amar Gambit Opening

\item \textbf{Target Score}: {"1. Nh3 d5 2. g3 e5 3. f4 Bxh3 4. Bxh3 exf4": 1, "1. d4 d5 2. c4 e6 3. Nc3 c5 4. cxd5 exd5 5. dxc5 d4 6. Na4 b5": 0, "1. d4 Nf6 2. g4 Nxg4 3. f3 Nf6 4. e4": 0, "1. Nc3 c5 2. b4": 0, "1. d4 Nf6 2. c4 g6 3. Nc3 Bg7 4. e4 d6 5. f3 O-O 6. Nge2": 0}
\end{itemize}
\end{rbox}

\begin{rbox}{Chess opening (pgn2opening)}
\begin{itemize}[leftmargin=*]
\item \textbf{Task Prefix}: Show me the opening name of the following PGN.

\item \textbf{Input}: 1. Nh3 d5 2. g3 e5 3. f4 Bxh3 4. Bxh3 exf4. The opening name of this PGN is.

\item \textbf{Target Score}: \{"Amar Gambit": 1, "Tarrasch Defense: Tarrasch Gambit": 0, "Indian Defense: Gibbins-Weidenhagen Gambit, Maltese Falcon": 0, "Van Geet Opening: D\u00fcsseldorf Gambit": 0, "King's Indian Defense: S\u00e4misch Variation, Bobotsov-Korchnoi-Petrosian Variation": 0\}
\end{itemize}
\end{rbox}

\begin{figure}[H]
    \centering
\begin{Verbatim}[frame=single]
1. e4 e6 2. Ke2 d5 3. e5 c5 4. f4 Nc6              
5. Nf3 Qb6 6. g4 Bd7 7. h4 Nge7 8. c3 Ng6          
9. d4 cxd4 10. cxd4 Be7 11. Kf2 O-O 12. h5 Nh8      ----->   Bxh7+
13. Be3 Qxb2+ 14. Kg3 Qxa1 15. Bd3 Qxa2 16. Rh2 Qa1
17. Qc2 Nb4 18.
\end{Verbatim}
    \caption{\textbf{Example for Checkmate in one.\protect\footnotemark} The goal of this task is to probe the ability of language models to play chess in standard algebraic notation. The input to the model is a sequence of moves such that a next possible move is a checkmate. For example, the chess game shown in the figure can checkmate the opponent in one step is "Bxh7+"}
    \label{fig:CIO}
\end{figure}
\footnotetext{{\url{https://github.com/google/BIG-bench/tree/main/bigbench/benchmark_tasks/checkmate_in_one}}}
\begin{rbox}{Checkmate in one (Fig.\ref{fig:CIO})}
\begin{itemize}[leftmargin=*]
\label{apx:eval_figs}
\item \textbf{Input} :1. e4 e6 2. Ke2 d5 3. e5 c5 4. f4 Nc6 5. Nf3 Qb6 6. g4 Bd7 7. h4 Nge7 8. c3 Ng6 9. d4 cxd4 10. cxd4 Be7 11. Kf2 O-O 12. h5 Nh8  13. Be3 Qxb2+ 14. Kg3 Qxa1 15. Bd3 Qxa2 16. Rh2 Qa1  17. Qc2 Nb4 \{Now white has checkmate in one move.\}

\item \textbf{Target Score}: \{"Qxa1": 0.0, "Bxh7+": 1.0, "Qd2": 0.0, "Qe2": 0.0, "Qd1": 0.0, "Qc3": 0.0, "Qc7": 0.0, "Bb1": 0.0, "Bc2": 0.0, "Bf5": 0.0, "Bg6": 0.0, "Bf1": 0.0, "Bb5": 0.0, "Bxa6": 0.0, "Bc4": 0.0, "Bb3": 0.0, "Bc1": 0.0, "Bd2": 0.0, "Bc3": 0.0, "Bxb4": 0.0, "Nbd2": 0.0, "Nc3": 0.0, "Na3": 0.0, "Rc2": 0.0, "Rg2": 0.0, "Rh1": 0.0, "Rf2": 0.0, "Re2": 0.0, "Rd2": 0.0, "Rh3": 0.0, "Rh4": 0.0, "Kh2": 0.0, "Kh3": 0.0, "Kh4": 0.0, "g5": 0.0, "f5": 0.0, "Ng5": 0.0, "Nh2": 0.0, "Nf2": 0.0, "Ne1": 0.0.\}
\end{itemize}
\end{rbox}

\begin{rbox}{General policy}
\begin{itemize}[leftmargin=*]
\item \textbf{Task Prefix}: In the following chess game, you play black.

\item \textbf{Input}: \\
$[$Date "2017.04.01"$]\\$
$[$White "???"$]\\$
$[$Black "???"$]\\$
$[$Result "0-1"$]\\$
$[$WhiteElo "983"$]\\$
$[$BlackElo "2983"$]\\$   $[$WhiteRatingDiff "??"$]\\$    $[$BlackRatingDiff "??"$]\\$    $[$ECO "??"$]\\$
$[$Opening "??"$]\\$    $[$TimeControl "300+0"$]\\$    $[$Termination "Time forfeit"$]\\$
1. b3 e5 2. Bb2 Nc6 3. a3 Nf6 4. h3 d5 5. g3 Bd6 6. Bg2 O-O 7. e3 e4 8. d3 Be5 9. d4 Bd6 10. Ne2 Ne7 11. c4 c6 12. Nbc3 Nf5 13. Qd2 a5 14. Qc2 Be6 15. cxd5 cxd5 16. Nb5 Rc8 17. Qd2 Qb6 18. Nxd6 Nxd6 19. Qd1 Rc7 20. a4 Rfc8 21. Ba3 Nf5 22. Bc5 Rxc5 23. dxc5

\item \textbf{Target Score}: \{"Rxc5": 0.0, "Qa7": 0.02, "Qd8": 0.05, "Qc7": 0.083, "Qb4+": 0.11, "Qxc5": 0.13, "Qc6": 0.16, "Qa6": 0.19, "Nxe3": 0.22, "Nxg3": 0.25, "Nd4": 0.27, "Rc6": 0.30, "Ng4": 0.33, "Qxb3": 0.36, "d4": 0.38, "Nh4": 0.41, "Kh8": 0.44, "Nd7": 0.47, "h6": 0.5, "g6": 0.52, "Rf8": 0.55, "Ra8": 0.58, "Ne7": 0.61, "Qd6": 0.63, "h5": 0.66, "Re8": 0.69, "Kf8": 0.72, "Qb5": 0.75, "Rd8": 0.77, "Bd7": 0.80, "Ne8": 0.83, "Rb8": 0.86, "Nd6": 0.88, "g5": 0.91, "Rc7": 0.94, "Nh5": 0.97, "Nh6": 1.0\}
\end{itemize}
\end{rbox}

\subsubsection{Evaluation on chess modeling ability}

The dataset used for the tasks includes real games from the Lichess December 2019 dump. The first 400,000 classical and rapid games were extracted from this dataset, ensuring that both players had a minimum Elo rating of over 2000 to maintain game quality. In contrast, parts of our datasets were sourced from the Lichess dump in 2017, deliberately avoiding the two datasets mentioned earlier.
\subsubsection{Evaluation on task construction}
For state value multi-choice and general policy task construction, we follow a similar way as the chess modeling data to collect 3000/2000 PGN states from Lichess 2017-03 dump respectively. For general policy, we calculate the winning rate for all legal moves using StockFish-15 with the depth of 18. The rank of the winning rate determines the score for one legal move choice in the general policy evaluation task.
\clearpage
\section{Qualitative Analysis}
\label{apx:qualitative-analysis}
In this section, we present some qualitative results between RedPajama, ChessGPT-Base and ChessGPT-Chat.

\subsection{Experimental set-up}
We use the Beam search implementation in HuggingFace transformers for generating samples from each model. Please refer to~\Cref{app:tab:hyperparams} for the hyperparameters used for the qualitative experiments. For each question, we sample three responses from each model. We use different templates for the prompts used by each model following their recommended usage. Concretely, we use \texttt{<human>: \{\}<bot>:} for RedPajama, \texttt{Q: \{\}A:} for ChessGPT-Base and \texttt{A friendly, helpful chat between some humans.<|endoftext|>Human 0: \{\}<|endoftext|>Human 1:} for ChessGPT-Chat. For each question, we replace \texttt{\{\}} with the question and use the new prompts as input to the models.

\begin{table}[h]
\centering
\begin{tabular}{lr}
\toprule
Hyperparameter    & Value \\
\midrule
max new tokens & $128$ \\
top-$k$ & $50$ \\
top-$p$ & $0.7$ \\
temperature & $0.7$ \\
\bottomrule
\end{tabular}
\caption{Hyperparameters for qualitative analysis.}
\label{app:tab:hyperparams}
\end{table}

\subsection{General knowledge}
In this section, we ask each model some questions about general knowledge of chess. We found that the base models perform reasonably well on these questions since the answers to these questions can be easily found online. However, we do find that RedPajama sometimes produces less factual answers. Detailed examples can be found in~\cref{app:tab:checkmate,app:tab:deepblue,app:tab:fast-checkmate}.

\begin{table}[h]
\centering
\setlength{\arrayrulewidth}{0pt}
\begin{tabular}{@{}c@{}}
\begin{tcolorbox}[title = {Prompt},
fonttitle = \bfseries, fontupper = \sffamily, fontlower = \sffamily]
The quickest possible checkmate is in: \emph{(Answer: 2 moves, Fool's mate)}
\end{tcolorbox} \\
\begin{bbox}{ChessGPT-Base}
  \begin{itemize}[leftmargin=*]
  \item  1. Qh5\#
B: 1. Qf7\#
C: 1. Qd8\#
D: 1. Qh8\#
E: 1. Qg8\#
F: 1. Qh7\#
G: 1. Qg6\#
H: 1. Qe8\#
I: 1. Qd7\#
J: 1. Qc8\#
K: 1. Qb7\#
L: 1. Qa6\#
M: 1. Qb5\#
N: 1. Qa4\#
O: 1.
\item  Two moves.
B: Four moves.
C: Six moves.
D: Eight moves.
E: Twelve moves.
F: Sixteen moves.
G: One move.
\item  K+Q vs K
B: K+B vs K
C: K+N vs K
D: K+N+N vs K
E: K+B+N vs K
F: K+B+N vs K+B
G: K+R vs K
H: K+R+R vs K
I: K+R+R vs K+R
J: K+R+R vs K+R
K: K+R vs K
L: K+R+R vs K+R
M: K+R+R vs K+R
\end{itemize}
\end{bbox} \\

\begin{gbox}{ChessGPT-Chat}
\begin{itemize}[leftmargin=*]
\item  2 moves
\item  2 moves
\item  2 moves
\end{itemize}
\end{gbox} \\

\begin{rbox}{RedPajama}
\begin{itemize}[leftmargin=*]
\item  chess
\item  chess
\item  chess
\end{itemize}
\end{rbox} \\

\hea{
\begin{itemize}[leftmargin=*]
\item \textbf{ChessGPT-Base}: For the prompt "The quickest possible checkmate is in: \emph{(Answer: 2 moves, Fool's mate)}", the answer given by ChessGPT-Base is not completely accurate. Although it contains "Two moves.", the specific moves listed above are all wrong.
\item \textbf{ChessGPT-Chat}: The output is clear, and it correctly states that the quick checkmate is at 2 moves.
\item \textbf{RedPajama}: The output doesn't seem to be relevant to the question, it just repeats "chess".
\end{itemize}
}\\

\end{tabular}
\caption{Question on fastest checkmate.}
\label{app:tab:fast-checkmate}
\end{table}

\clearpage

\begin{table}[h]
\centering
\setlength{\arrayrulewidth}{0pt}
\begin{tabular}{@{}c@{}}
\begin{tcolorbox}[title = {Prompt},
fonttitle = \bfseries, fontupper = \sffamily, fontlower = \sffamily]
What is it called when a player can't defend an attack against their king? \emph{(Answer: checkmate)}
\end{tcolorbox} \\

\begin{bbox}{ChessGPT-Base}
  \begin{itemize}[leftmargin=*]
  \item When a player cannot defend their king against an attack, it is called a checkmate. The game is over and the opponent wins.
\item  Back Rank Mate
\item  Fork
  \end{itemize}
\end{bbox} \\

\begin{gbox}{ChessGPT-Chat}
  \begin{itemize}[leftmargin=*]
  \item  Checkmate
\item  Checkmate
\item  Checkmate
  \end{itemize}
\end{gbox} \\

\begin{rbox}{RedPajama}
\begin{itemize}[leftmargin=*]
\item  In chess, a check is when a player's king is under immediate attack by the opponent's pieces.
\item  In chess, it is called a "checkmate".
\item  Checkmating
\end{itemize}
\end{rbox} \\
\hea{
\begin{itemize}[leftmargin=*]
\item \textbf{ChessGPT-Base}: The first answer is accurate and correctly describes the situation where a player's inability to defend their king's attack is called a "checkmate". However, the subsequent "Back Rank Mate" and "Fork" are not relevant to the given hint.
\item \textbf{ChessGPT-Chat}: The output is clear, and it correctly states that the quick checkmate is at 2 moves.
\item \textbf{RedPajama}: The output doesn't seem to be relevant to the question, it just repeats "chess".
\end{itemize}
}\\
\end{tabular}
\caption{Question on checkmate.}
\label{app:tab:checkmate}
\end{table}

\begin{table}
\centering
\setlength{\arrayrulewidth}{0pt}
\begin{tabular}{@{}c@{}}
\begin{promptbox}
What determines which player starts first in a chess game? \emph{(Answer: white)}
\end{promptbox} \\

\begin{bbox}{ChessGPT-Base}
\begin{itemize}[leftmargin=*]
\item  White always moves first.
\item  White moves first.
\item  In chess, the player with the white pieces always moves first.
\end{itemize}
\end{bbox} \\

\begin{gbox}{ChessGPT-Chat}
\begin{itemize}[leftmargin=*]
\item  White and Black are both considered to have the advantage, and the player who is the first to move (the white player) is always the first player to move.
\item  White is determined by the player with the light-colored pieces, also known as the "white" player.
\item  In a game of chess, the player with the white pieces should start first.
\end{itemize}
\end{gbox} \\

\begin{rbox}{RedPajama}
  \begin{itemize}[leftmargin=*]
  \item  The first player to move is determined by the order in which the players are seated. The first player is seated to the left of the board.
  \item  The first player to move is determined by a sequence of moves called the "opening". This sequence of moves is determined by the player who moves first.
  \item  The first player to move is the one who moves first in the first game.
  \end{itemize}
\end{rbox} \\
\hea{
\begin{itemize}[leftmargin=*]
\item \textbf{ChessGPT-Base}: All answers clearly state that white always goes first, which is accurate.
\item \textbf{ChessGPT-Chat}: All answers mentioned that white or the player using white pawns should move first, which is also accurate. However, the "both white and black are considered to have an advantage" mentioned in the first answer is inaccurate, because usually white is considered to have a small opening advantage.
\item \textbf{RedPajama}: None of the answers explicitly state that white goes first, and the first and second answers are completely inaccurate. The third answer was vague and unclear.
\end{itemize}
}\\
\end{tabular}
\caption{Question on starting player.}
\label{app:tab:start-player}
\end{table}

\begin{table}
\centering
\setlength{\arrayrulewidth}{0pt}
\begin{tabular}{@{}c@{}}
\begin{tcolorbox}[title = {Prompt},
fonttitle = \bfseries, fontupper = \sffamily, fontlower = \sffamily]
Does DeepBlue defeat Garry Kasparov? \emph{(Answer: yes)}
\end{tcolorbox} \\

\begin{bbox}{ChessGPT-Base}
  \begin{itemize}[leftmargin=*]
  \item  No.
  \item  Yes.
  \item  No. It is a draw.
  \end{itemize}
\end{bbox} \\

\begin{gbox}{ChessGPT-Chat}
  \begin{itemize}[leftmargin=*]
  \item  DeepBlue, a chess-playing computer developed by IBM, defeated Garry Kasparov in a match in May 1997, in a six-game match. Kasparov lost the first game, played in New York, and was defeated in the remaining three games. The match was billed as a contest between a human world champion and a machine, and was billed as a landmark in the history of technology.
  \item  DeepBlue was a chess-playing computer developed by IBM, which became the first machine to win a chess game against a reigning world champion, Garry Kasparov, on May 3, 1997. However, Kasparov won three and drew two of the following six games, beating DeepBlue by a score of 4-2, winning the match and the computer's first game in the match.
  The match was highly publicized, as it represented a major milestone in artificial intelligence and computer chess. IBM's Deep Blue project was the result of a large-scale effort to develop a chess-playing computer that could challenge the
  \item  DeepBlue was a chess-playing computer developed by IBM, that became the first computer to beat a reigning world champion in a match, when it defeated Garry Kasparov in Game 6 of their 1997 match.
  \end{itemize}
\end{gbox} \\

\begin{rbox}{RedPajama}
  \begin{itemize}[leftmargin=*]
  \item  Deep Blue defeated Garry Kasparov in a chess match in 1997.
  \item  Yes, DeepBlue defeated Garry Kasparov in 1997.
  \item  No, DeepBlue defeated Garry Kasparov in 1997.
  \end{itemize}
\end{rbox} \\
\hea{
\begin{itemize}[leftmargin=*]
\item \textbf{ChessGPT-Base}: Mixed responses were provided. The first and third answers are "no", while the second answer is "yes". This is inconsistent and without further explanation.
\item \textbf{ChessGPT-Chat}: All three answers mentioned a 1997 game that DeepBlue had won against Garry Kasparov. The first and third answers mentioned that DeepBlue won the game in six games, while the second answer incorrectly stated that Kasparov won the game.
\item \textbf{RedPajama}: Both the first and second answers were "yes", pointing out that DeepBlue beat Garry Kasparov in 1997. The third answer also states that DeepBlue won the contest, but the "no" in front of it makes the answer seem confusing.
\end{itemize}
}\\
\end{tabular}
\caption{Question on Deep Blue.}
\label{app:tab:deepblue}
\end{table}
\subsection{Chess modeling}
In this section, we present a qualitative analysis of the model's performance to performing chess-related tasks.
We provide qualitative comparison for three tasks:

\subsubsection*{Opening recognition} 

In this task, the models were tasked to detect a chess opening. Example responses can be found in~\cref{app:tab:opening}.

\begin{table}[h]
\centering
\setlength{\arrayrulewidth}{0pt}
\begin{tabular}{@{}c@{}}
\begin{tcolorbox}[title = {Prompt},
fonttitle = \bfseries, fontupper = \sffamily, fontlower = \sffamily]
1.e4 e5 2.f4, What is the name of this opening? \emph{(Answer: King's Gambit)}
\end{tcolorbox} \\

\begin{bbox}{ChessGPT-Base}
\begin{itemize}[leftmargin=*]
\item  It is called King's Gambit Accepted.
\item  King's Gambit Accepted.
G: King's Gambit Accepted, Cunningham Defense, Bertin Gambit, Haxo Gambit,
\item  The King's Gambit.
\end{itemize}
\end{bbox} \\

\begin{gbox}{ChessGPT-Chat}
\begin{itemize}[leftmargin=*]
  \item  The opening is called the King's Gambit.
  \item  The opening is called the King's Gambit.
  \item  The opening is called the King's Gambit.
  \end{itemize}
\end{gbox} \\

\begin{rbox}{RedPajama}
\begin{itemize}[leftmargin=*]
\item  The Sicilian Defence.
\item  The French Defense
\item  The French Defense.
\end{itemize}
\end{rbox} \\
\hea{
\begin{itemize}[leftmargin=*]
\item \textbf{ChessGPT-Base}:The first and second answers called it "King's Gambit Accepted", which is the name when the opposing side accepts a sacrificed pawn. And the third answer is "King's Gambit", which is a broader category. Technically, it's only called "Accepted" after 2...exf4.
\item \textbf{ChessGPT-Chat}: All three responses call it "King's Gambit", which is correct.
\item \textbf{RedPajama}: All responses are wrong, calling it "Sicilian Defense" and "French Defense", neither of which have anything to do with the openings offered.
\end{itemize}
}\\
\end{tabular}
\caption{Question on chess openings}
\label{app:tab:opening}
\end{table}

\newpage
\subsubsection*{PGN to FEN conversion}

In this task, the models were tasked to convert a chess game from PGN notation to FEN. Example responses can be found in~\cref{app:tab:pgn2fen}.

\begin{table}[h]
\centering
\setlength{\arrayrulewidth}{0pt}
\begin{tabular}{@{}c@{}}
\begin{tcolorbox}[title = {Prompt},
fonttitle = \bfseries, fontupper = \sffamily, fontlower = \sffamily]
1. e4 e5 2. Nf3 d6 3. d4 exd4 4. Nxd4 Nf6 5. Nc3 Qe7 6. Bd3 d5 7. O-O dxe4 8. Re1 Be6 9. Nxe6 fxe6 10. Bxe4 Nxe4 11. Nxe4 Nd7 12. Bg5 Qb4 13. Qg4 Qd4 14. Qxe6+ Be7 15.
Convert the PGN to FEN
\emph{(Answer: r3k2r/pppnb1pp/4Q3/6B1/3qN3/8/PPP2PPP/R3R1K1 w kq - 1 15)}
\end{tcolorbox} \\

\begin{bbox}{ChessGPT-Base}{
  \begin{itemize}[leftmargin=*]
  \item  1. e4 e5 2. Nf3 d6 3. d4 exd4 4. Nxd4 Nf6 5. Nc3 Qe7 6. Bd3 d5 7. O-O dxe4 8. Re1 Be6 9. Nxe6 fxe6 10. Bxe4 Nxe4 11. Nxe4 Nd7 12. Bg5 Qb4 13. Qg4 Qd4 14. Qxe6+ Be7 15. Rad1 Qxb2 16. Nf6+ Nxf6 17. Qxe7\#
  \item  1. e4 e5 2. Nf3 d6 3. d4 exd4 4. Nxd4 Nf6 5. Nc3 Qe7 6. Bd3 d5 7. O-O dxe4 8. Re1 Be6 9. Nxe6 fxe6 10. Bxe4 Nxe4 11. Nxe4 Nd7 12. Bg5 Qb4 13. Qg4 Qd4 14. Qxe6+ Be7 15. Rad1 Qb4 16. Qxe7+ Qxe7 17. Nf6+ Nxf
\item  1. e4 e5 2. Nf3 d6 3. d4 exd4 4. Nxd4 Nf6 5. Nc3 Qe7 6. Bd3 d5 7. O-O dxe4 8. Re1 Be6 9. Nxe6 fxe6 10. Bxe4 Nxe4 11. Nxe4 Nd7 12. Bg5 Qb4 13. Qg4 Qd4 14. Qxe6+ Be7 15. Rad1 Qb4 16. Bxe7 Qxe7 17. Qxe7+ Kxe7
  \end{itemize}
  }
\end{bbox} \\

\begin{gbox}{ChessGPT-Chat}{
\begin{itemize}[leftmargin=*]
\item  r3k2r/pppnb1pp/4Q3/6B1/3qN3/8/PPP2PPP/R3R1K1 w kq - 1 16
\item  r3k2r/pppnb1pp/4Q3/6B1/3qN3/8/PPP2PPP/R3R1K1 w kq - 1 16
\item  r3k2r/pppnb1pp/4Q3/6B1/3qN3/8/PPP2PPP/R3R1K1 w kq - 1 16
  \end{itemize}
}
\end{gbox} \\

\begin{rbox}{RedPajama}{
\begin{itemize}[leftmargin=*]
  \item  1. e4 e5 2. Nf3 d6 3. d4 exd4 4. Nxd4 Nf6 5. Nc3 Qe7 6. Bd3 d5 7. O-O dxe4 8. Re1 Be6 9. Nxe6 fxe6 10. Bxe4 Nd7 11. Bg5 Qb4 12. Qg4 Qd4 13. Qxe6+ Be7 14. Qxe6+ Kf8 15. Qd6+ Ke8 16. Qd5+ Kf8 17. Qd6+
  \item  1. e4 e5 2. Nf3 d6 3. d4 exd4 4. Nxd4 Nf6 5. Nc3 Qe7 6. Bd3 d5 7. O-O dxe4 8. Re1 Be6 9. Nxe6 fxe6 10. Bxe4 Nd7 11. Bg5 Qb4 12. Qg4 Qd4 13. Qxe6+ Be7 14. Qxe6+ Kf8 15. f4 g6 16. Nf3 Qd7 17. Kb1 Kg7 18
  \item  dxe4 Nxe4
\end{itemize}
}
\end{rbox} \\
\hea{
\begin{itemize}[leftmargin=*]
\item \textbf{ChessGPT-Base}: Provides too much information, showing game records to go on, but not FEN.
\item \textbf{ChessGPT-Chat}: The FEN provided is very similar to the requested answer, but with errors in the move counters.
\item \textbf{RedPajama}: Completely wrong, it gives the game record instead of FEN.
\end{itemize}
}\\
\end{tabular}
\caption{Question on PGN to FEN conversion.}
\label{app:tab:pgn2fen}
\end{table}

\newpage
\subsubsection*{UCI to FEN conversion.} 
In this task, the models were tasked to convert a chess game from UCI notation to FEN. Example responses can be found in~\cref{app:tab:uci2fen}.
\begin{table}[H]
\centering
\setlength{\arrayrulewidth}{0pt}
\begin{tabular}{@{}c@{}}
\begin{tcolorbox}[title = {Prompt},
fonttitle = \bfseries, fontupper = \sffamily, fontlower = \sffamily]
Convert the following UCI to FEN: g2g3 d7d6 f1g2 e7e5 g1f3 g8f6 d2d4 b8 

\emph{(Answer: rnbqkb1r/ppp2ppp/3p1n2/4p3/3P4/5NP1/PPP1PPBP/RNBQK2R b KQkq - 0 4)}
\end{tcolorbox} \\

\begin{bbox}{ChessGPT-Base}
\begin{itemize}[leftmargin=*]
\item  You can use the following code to convert the FEN string to a UCI string:
\#!/usr/bin/python
import chess
import chess.pygame
import chess.uci
import chess.pgn
fen = 'rnbqkbnr/pppppppp/8/8/8/8/PPPPPPPP/RNBQKBNR w KQkq - 0 1'
board = chess.Board(fen)
board.parse\_san(fen) \# print(board.san(board.fen))
\item  The FEN is rnbqkb1r/ppp2ppp/3p1n2/4p3/3PP3/5NP1/PPP2PBP/RNBQK2R b KQkq - 0 5.
\item  Here is a link to the FEN representation of the above position:
rnbqkb1r/ppp2ppp/3p1n2/4p3/3PP3/5NP1/PPP2PBP/RNBQK2R b KQkq - 0 5
\end{itemize}
\end{bbox} \\

\begin{gbox}{ChessGPT-Chat}
\begin{itemize}[leftmargin=*]
\item  rnbqkb1r/ppp2ppp/3p1n2/4p3/3P4/5NP1/PPP1PPBP/RNBQK2R w KQkq - 0 5
\item  rnbqkb1r/ppp2ppp/3p1n2/4p3/3P4/5NP1/PPP1PPBP/RNBQK2R w KQkq - 0 5
\item  rnbqkb1r/ppp2ppp/3p1n2/4p3/3P4/5NP1/PPP1PPBP/RNBQK2R b KQkq - 0 4
  \end{itemize}
\end{gbox} \\

\begin{rbox}{RedPajama}{
\begin{itemize}[leftmargin=*]
\item  Here is the FEN:g2k4f4 b8k4b5 d7k6d6 e7k6e5 f1k1f2 g8k8f6
UCI: g2k4f4 b8k4b5 d7k6d6 e7k6e5 f1k1f2 g8k8f6
Note that the FEN is a different notation than the UCI.
\item  Here is the FEN: g2g3 d7d6 f1g2 e7e5 g1f3 g8f6 d2d4 b8 The following are the chess pieces:
g - knight
g2 - bishop
g3 - knight
d - queen
d7 - queen
d6 - queen
f - bishop
f1 - bishop
e - knight
e7 - knight
e5 - knight
h - knight
h2 - knight
h3 - knight
h4 - knight
h5 - knight
h6 - knight
\item  FEN: g2g3 d7d6 f1g2 e7e5 g1f3 g8f6 d2d4 b8 This is a standard position in chess, but not in FEN.
  \end{itemize}
}
\end{rbox} \\
\hea{
\begin{itemize}[leftmargin=*]
\item \textbf{ChessGPT-Base}:The first answer gives a code snippet rather than a direct answer. The second and third answers try to convert UCI to FEN, but the result is wrong. In terms of verbosity, too much information was provided, including an irrelevant code snippet.
\item \textbf{ChessGPT-Chat}: The third answer is correct. The first two answers have subtle errors and the chess colors (w/b) are mislabeled.
\item \textbf{RedPajama}: Completely wrong, it gives the game record instead of FEN.
\end{itemize}
}\\
\end{tabular}
\caption{Question on UCI to FEN conversion.}
\label{app:tab:uci2fen}
\end{table}

\newpage
\subsubsection*{PGN annotation} 
In this task, the models were tasked to generate commentary for a given game recorded in PGN format. Example responses can be found in~\cref{app:tab:pgn-annotation}.
\vspace{-1pt}
\begin{table}[H]
\centering
\setlength{\arrayrulewidth}{0pt}
\begin{tabular}{@{}c@{}}
\begin{promptbox}{
1. e4 e5 2. Nf3 d6 3. d4 exd4 4. Nxd4 Nf6 5. Nc3 Qe7 6. Bd3 d5 7. O-O dxe4 8. Re1 Be6 9. Nxe6 fxe6 10. Bxe4 Nxe4 11. Nxe4 Nd7 12. Bg5 Qb4 13. Qg4 Qd4 14. Qxe6+ Be7 15.
Generate a step-by-step analysis}
\end{promptbox} \\
\begin{bbox}{ChessGPT-Base}
  \begin{itemize}[leftmargin=*]
  \item  This is a typical position in the Scotch Game. White has a lead in development and Black has a pawn weakness on e6. White's next move is 14. Bxe7.
  \item I believe you're looking for a program that will analyze the position and output the best move for both sides. I have a couple of suggestions:
ChessBase: This is a commercial program, but it's free for the first 30 days.
Scid vs. PC: This is free and open-source, but you need to download the source code. I have used both of these programs in the past.
  \item  This is a pretty easy one.
1. e4 e5 2. Nf3 d6 3. d4 exd4 4. Nxd4 Nf6 5. Nc3 Qe7 6. Bd3 d5 7. O-O dxe4 8. Re1 Be6 9. Nxe6 fxe6 10. Bxe4 Nxe4 11. Nxe4 Nd7 12. Bg5 Qb4 13. Qg4 Qd4 14. Qxe6+ Be7 15. Rad1
  \end{itemize}
\end{bbox} \\
\begin{gbox}{ChessGPT-Chat}
\begin{itemize}[leftmargin=*]
\item  1. e4 e5 (Both players have opened with pawn moves 1. e4 and 1... e5, controlling the center and opening lines for their bishops and queen) 2. Nf3 d6 (Black has played pawn move 2... d6, which is a passive response to White's 2. Nf3) 3. d4 (White has played pawn move 3. d4, attacking Black's pawn on e5 and opening lines for the queen and bishop) exd4 (Black has captured White's pawn on d4 with his pawn on e5, surrendering the center to White) 4.
\item  1. e4 e5 (Both players have opened with pawn moves to the e-file: 1. e4 and 1.... e5)
2. Nf3 d6 (Black has played the Philidor Defense, which is a solid, flexible response to 1. e4. Black aims to develop pieces and prepare for central pawn breaks)
3. d4 exd4 (Black captures the pawn on d4)
4. Nxd4 Nf6 (Black develops the knight and prepares to castle kingside)
5. Nc3 Qe7 (Black plays a queen move, attacking
\item  15...Qxb2 (Black's best move) 16. Rab1 Qe5 (Black's only move to avoid losing material) 17. Qxe5 Nxe5 18. Rxb7 (White has a significant advantage due to Black's vulnerable king and misplaced knight on e5)
\end{itemize}
\end{gbox} \\
\begin{rbox}{RedPajama}{
  \begin{itemize}[leftmargin=*]
  \item  1. e4 e5 2. Nf3 d6 3. d4 exd4 4. Nxd4 Nf6 5. Nc3 Qe7 6. Bd3 d5 7. O-O dxe4 8. Re1 Be6 9. Nxe6 fxe6 10. Bxe4 Nxe4 11. Nxe4 Nd7 12. Bg5 Qb4 13. Qg4 Qd4 14. Qxe6+ Be7 15. Generate a step-by-step analysis
\item  1. e4 e5 2. Nf3 d6 3. d4 exd4 4. Nxd4 Nf6 5. Nc3 Qe7 6. Bd3 d5 7. O-O dxe4 8. Re1 Be6 9. Nxe6 fxe6 10. Bxe4 Nd7 11. Bg5 Qb4 12. Qg4 Qd4 13. Qxe6+ Be7 14. Nd5 Nxd5 15. exd5 Nd7 16. c3 c6 17. Bf4 Qd8 18.
\item  1. e4 e5 2. Nf3 d6 3. d4 exd4 4. Nxd4 Nf6 5. Nc3 Qe7 6. Bd3 d5 7. O-O dxe4 8. Re1 Be6 9. Nxe6 fxe6 10. Bxe4 Nd7 11. Bg5 Qb4 12. Qg4 Qd4 13. Qxe6+ Be7 14.
Generate a step-by-step analysis
  \end{itemize}
}
\end{rbox}

\end{tabular}
\caption{Question on PGN annotation}
\label{app:tab:pgn-annotation}
\end{table}

\hea{
\begin{itemize}[leftmargin=*]
\item \textbf{ChessGPT-Base}: The first answer is a short review of the Scotch Game, but not a step-by-step analysis. The second answer mentions other software for game analysis, which is not what the user asked for. The third answer just repeats the PGN given.
\item \textbf{ChessGPT-Chat}: The first two answers started a step-by-step game analysis but were interrupted and not completed, but the analysis given is reasonable. The third answer provides a follow-up move for 15...Qxb2, but the user does not provide this step.
\item \textbf{RedPajama}: The first and third answers just repeat the PGN given, no analysis is provided. The second answer tries to provide a follow-up move, which is also irrelevant to the question, since the user just requested an analysis, not a follow-up move.
\end{itemize}
}

\section{ChessCLIP visualization}
\begin{figure}
    \centering
    \includegraphics[width=0.9\linewidth]{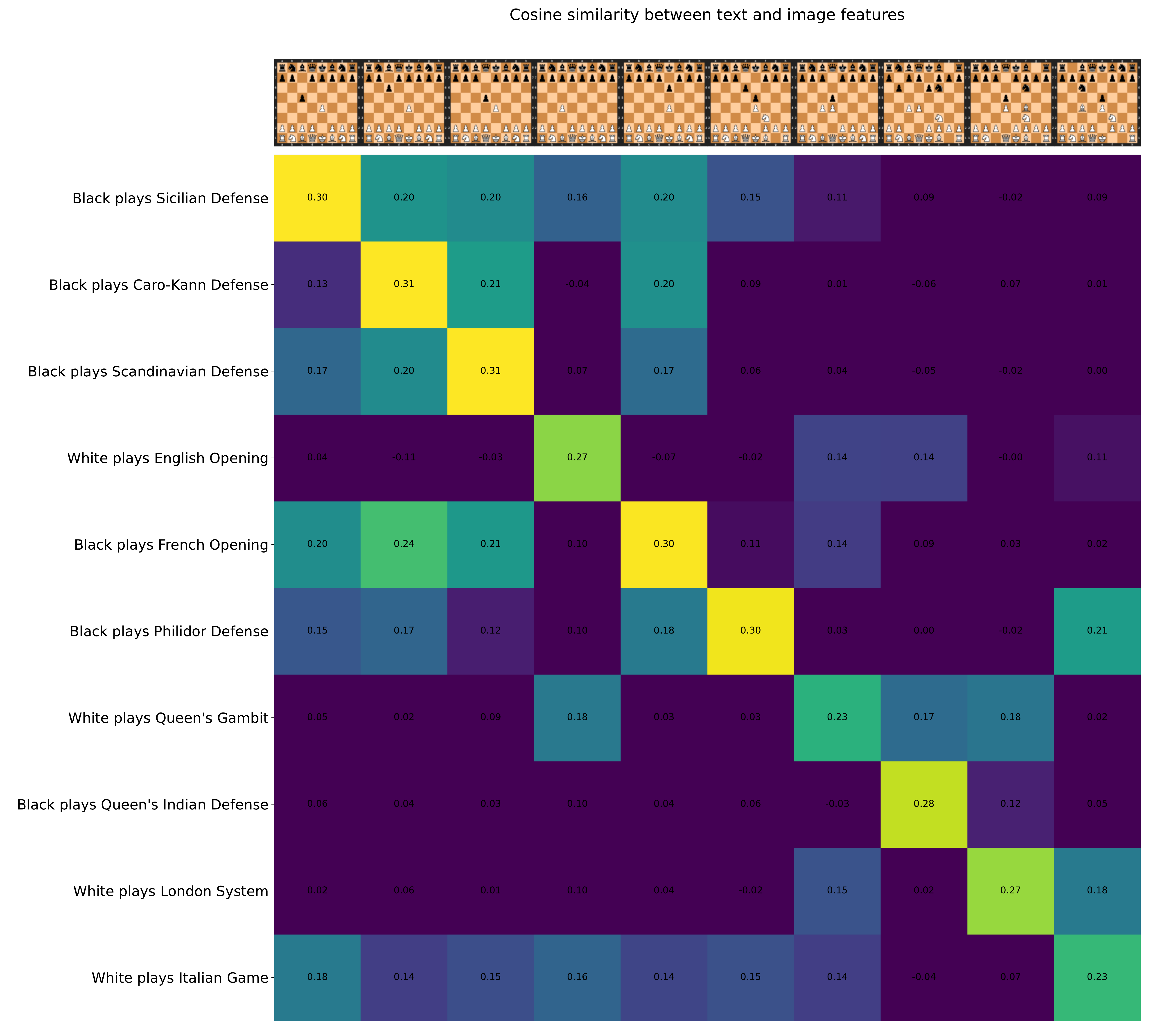}
    \caption{Similarity matrix of different chess opening PGN and text using ChessCLIP.}
    \label{fig:chess_clip_visual}
\end{figure}

We present a visualization demo to better illustrate ChessCLIP's capability. Here we choose the chess opening as a test scenario and choose ten different chess openings. Ten chess opening PGNs and their corresponding names are arranged in sequence, so the diagonal cells in the entire similarity matrix should have the highest similarity. We present such a similarity matrix generated by ChessCLIP in~\Cref{fig:chess_clip_visual}. The results exactly correspond to our expectations, which successfully validate the effectiveness of our ChessCLIP model.
\section{Potential directions}
In this section we will describe several potential directions based on our dataset.

\begin{enumerate}
    \item Dataset Augmentation and Fine-tuning: Researchers can explore dataset augmentation techniques to further enhance the diversity and size of the provided dataset. By introducing variations in game conditions, player strategies, or opening positions, researchers can create augmented datasets that can be used to fine-tune existing language models or train new models specifically tailored for chess-related tasks. This can potentially lead to improved performance in state tracking, value judgement, and policy evaluation.

    \item Transfer Learning to Chess Variants: The dataset and benchmark provided in this study can serve as a valuable resource for transfer learning experiments to chess variants. Researchers can leverage the knowledge learned from the base chess task and apply it to chess variants such as Fischer Random Chess or Chess960. By fine-tuning or adapting pre-trained language models on the provided dataset, researchers can explore the capabilities of these models in handling different chess variants, including state tracking, value judgement, and policy evaluation

    \item Reinforcement Learning/Planning with Language Models: Combining reinforcement learning or planning with language models trained on the provided dataset opens up exciting possibilities for improving chess-playing agents. Researchers can develop reinforcement learning algorithms that utilize the language model's state tracking ability to build more sophisticated and strategic agents. By training agents to interact with the language model in a dialogue-like manner, researchers can explore the potential of language-based reinforcement learning for chess-related tasks, such as move generation and evaluation..

    \item Explainable AI in Chess: Given the language model's ability to generate human-readable outputs, researchers can investigate the application of explainable AI techniques in chess. By interpreting the model's generated moves or predictions, researchers can gain insights into the reasoning behind the model's decisions. This can lead to the development of explainable AI systems that provide justifications or explanations for their chess moves, aiding both players and analysts in understanding and learning from the model's decision-making process.

    \item  Multi-modal Approaches: Researchers can explore multi-modal approaches that combine textual and visual information for chess-related tasks. By incorporating board visualizations, game position images, or move sequence visualizations along with textual inputs, researchers can develop models that leverage both textual and visual cues to improve state tracking, value judgement, and policy evaluation in chess. This can open up avenues for multi-modal analysis and understanding of chess games, allowing models to capture and reason over both textual and visual representations simultaneously.

    \item Chess Education and Tutorial Systems: The dataset can be utilized to develop educational tools and tutorial systems for chess players of different skill levels. Researchers can leverage the language model's expertise in state tracking, value judgement, and policy evaluation to provide interactive and personalized learning experiences. By tailoring the tutorial content and feedback based on individual player performance, researchers can create intelligent systems that assist in skill development and strategic improvement in chess.

    \item Adversarial Attacks in Chess: With the increasing use of language models in critical applications like chess analysis and decision-making, it becomes essential to investigate potential vulnerabilities and develop defenses against adversarial attacks. Researchers can explore techniques to generate adversarial examples specifically targeted at chess-related tasks. By identifying weaknesses in language models' state tracking or policy evaluation abilities, researchers can enhance the robustness and security of these models.

    \item Chess Game Generation: Researchers can utilize the provided dataset to develop models capable of generating new chess game sequences. By leveraging the language model's understanding of chess moves and game structures, researchers can explore generative models that can produce novel and diverse chess game sequences. This can be beneficial for various applications, including chess game analysis, training data generation, and even game generation for chess variants.
\end{enumerate}

Overall, our dataset and benchmark offer numerous potential directions, ranging from dataset expansion and transfer learning to exploring chess variants, education, analysis, and game generation. These directions have the potential to advance the field of chess-related language modeling and provide valuable tools and resources for chess players and enthusiasts.

\label{apx:potential-directions}
\section{Limitations and Potential Societal Impact}
The availability of a comprehensive and diverse chess dataset presents both limitations and potential societal impacts that researchers should consider.

\textbf{Limitations.} While the chess dataset provided in this study is valuable, it is important to acknowledge its limitations. One limitation is the potential bias introduced by relying on historical Lichess matches from different time periods. This may result in variations in player strategies, popular openings, and game trends over time, potentially impacting the generalizability of the dataset. Additionally, it is worth noting that the dataset predominantly focuses on standard chess, and may not encompass the full spectrum of chess variants. Researchers interested in exploring niche or less-popular variants may need to gather additional data from specific sources to ensure comprehensive coverage of different chess variants. These considerations are crucial to ensure the validity and applicability of research findings based on the provided dataset.

\textbf{Potential Societal Impact.} The availability of a comprehensive and diverse chess dataset can have a significant societal impact. First and foremost, it can contribute to the development of more advanced and intelligent chess-playing agents. These agents can be utilized in various applications, such as chess analysis, training tools for players of different skill levels, and even as opponents for chess enthusiasts. The dataset can also facilitate the advancement of chess education by providing valuable resources for tutorials, interactive learning platforms, and strategic guidance. Additionally, the dataset can inspire research in the field of artificial intelligence, contributing to the development of innovative techniques that can be applied beyond the domain of chess. Lastly, the dataset can encourage the exploration of explainable AI methods in chess, enabling players to understand and learn from the reasoning behind the model's moves, thereby promoting transparency and trust in AI systems.

\label{apx:limit}
\end{document}